\documentclass[10pt,twocolumn,letterpaper]{article}

\usepackage[pagenumbers]{cvpr} 

\usepackage[dvipsnames]{xcolor}

\usepackage{graphicx}
\usepackage{algorithm}
\usepackage{algpseudocode}
\usepackage[aboveskip=2pt]{subcaption}
\usepackage{makecell}
\usepackage{comment}

\DeclareMathOperator*{\argmin}{arg\,min}

\definecolor{cvprblue}{rgb}{0.21,0.49,0.74}
\usepackage[pagebackref,breaklinks,colorlinks,citecolor=cvprblue]{hyperref}

\usepackage{caption}
\usepackage[accsupp]{axessibility} 

\def\paperID{16428} 
\def\confName{CVPR}
\def\confYear{2024}

\title{Distilling ODE Solvers of Diffusion Models into Smaller Steps}

\author{Sanghwan Kim$^1$ \quad Hao Tang$^{1,2}$\thanks{Corresponding author} \quad Fisher Yu$^1$\\
$^1$ETH Zürich \quad $^2$Carnegie Mellon University\\
}

\begin{document}
\maketitle

\begin{abstract}

Diffusion models have recently gained prominence as a novel category of generative models. Despite their success, these models face a notable drawback in terms of slow sampling speeds, requiring a high number of function evaluations (NFE) in the order of hundreds or thousands. In response, both learning-free and learning-based sampling strategies have been explored to expedite the sampling process. Learning-free sampling employs various ordinary differential equation (ODE) solvers based on the formulation of diffusion ODEs. However, it encounters challenges in faithfully tracking the true sampling trajectory, particularly for small NFE. Conversely, learning-based sampling methods, such as knowledge distillation, demand extensive additional training, limiting their practical applicability. To overcome these limitations, we introduce \emph{Distilled-ODE solvers} (D-ODE solvers), a straightforward distillation approach grounded in ODE solver formulations. Our method seamlessly integrates the strengths of both learning-free and learning-based sampling.

D-ODE solvers are constructed by introducing a single parameter adjustment to existing ODE solvers. Furthermore, we optimize D-ODE solvers with smaller steps using knowledge distillation from ODE solvers with larger steps across a batch of samples. Comprehensive experiments demonstrate the superior performance of D-ODE solvers compared to existing ODE solvers, including DDIM, PNDM, DPM-Solver, DEIS, and EDM, particularly in scenarios with fewer NFE. Notably, our method incurs negligible computational overhead compared to previous distillation techniques, facilitating straightforward and rapid integration with existing samplers. Qualitative analysis reveals that D-ODE solvers not only enhance image quality but also faithfully follow the target ODE trajectory.

\end{abstract}

\section{Introduction}\label{sec:introduction}

Diffusion models~\citep{sohl2015deep, ho2020denoising, song2019generative} have recently emerged as a compelling framework for generative models, demonstrating state-of-the-art performance across diverse applications such as image generation~\citep{song2020score, dhariwal2021diffusion}, text generation~\citep{hoogeboom2021argmax, austin2021structured}, audio generation~\citep{mittal2021symbolic, lu2021study}, 3D shape generation~\citep{cai2020learning, luo2021diffusion}, video synthesis~\citep{harvey2022flexible, yang2022diffusion}, and graph generation~\citep{niu2020permutation, vignac2022digress}.


While diffusion models excel at producing high-quality samples and mitigating issues like mode collapse~\citep{salimans2016improved, zhao2018bias}, their sampling process often demands a substantial number of network evaluations, rendering the process slow and computationally intensive~\citep{xiao2021tackling}. Recent research has focused on optimizing the sampling process to enhance efficiency without compromising sample quality~\citep{song2020denoising, karras2022elucidating, salimans2022progressive}. Notably, methods targeting improved sampling efficiency within diffusion models fall into two main categories: \textit{learning-free sampling} and \textit{learning-based sampling}~\citep{yang2022diffusion2}.

Learning-free sampling can be applied to pre-trained diffusion models without additional training and often involves efficient solvers for stochastic differential equations (SDEs) or ordinary differential equations (ODEs)~\citep{song2020score}. Notable examples include DDIM~\citep{song2020denoising}, which employs a non-Markovian process for accelerated sampling, and PNDM~\citep{liu2021pseudo}, introducing a pseudo-numerical method for solving differential equations on given data manifolds. EDM~\citep{karras2022elucidating} utilizes Heun's second-order method, demonstrating improved sampling quality over naive Euler's method~\citep{song2020score}. Recently, DPM-Solver~\cite{lu2022dpm} and DEIS~\cite{zhang2022fast} leverage the semi-linear structure of diffusion ODEs and employ numerical methods of exponential integrators. These ODE solvers aim for accurate score function estimation along the ODE sampling trajectory where the density of data distribution is high~\cite{liu2021pseudo, zheng2023fast}. However, the sampling path of ODE solvers may deviate from the true trajectory, especially with a small number of denoising steps, resulting in significant fitting errors in the score function~\cite{li2023alleviating, xiao2021tackling, song2023consistency}.

In contrast, learning-based sampling involves additional training to optimize specific objectives, such as knowledge distillation~\citep{salimans2022progressive, song2023consistency} and optimized discretization~\citep{nichol2021improved, watson2021learning}. For instance, progressive distillation~\citep{salimans2022progressive} iteratively distills pre-trained diffusion models into a student model requiring fewer sampling steps. Recently, \citet{song2023consistency} introduced consistency models, trained to predict consistent outputs along the same ODE trajectory. 
Although distillation-based techniques enable generation within a few steps, extensive training is needed to adapt the denoising network of the diffusion models to each dataset, sampler, and network.

To address these challenges, we propose a novel distillation method for diffusion models called Distilled-ODE solvers (D-ODE solvers), leveraging inherent sampling dynamics in existing ODE solvers. D-ODE solvers bridge the gap between learning-free and learning-based sampling while mitigating associated issues. Our approach is grounded in the observation that the outputs of the denoising network (\ie, denoising output) exhibit a high correlation within neighboring time steps.

D-ODE solvers introduce a single additional parameter to ODE solvers, linearly combining the current denoising network output with the previous one. This allows a more accurate estimation of the denoising output at each timestep $t$. For high-order solvers (\eg, PNDM, DEIS, and DPM-Solver), we linearly combine their high-order estimations to leverage their ability to approximate the true score function. The additional parameter is optimized for each dataset by minimizing the difference between the output of D-ODE solvers with smaller steps (student) and that of ODE solvers with larger steps (teacher). Once the optimal parameter is established, D-ODE solvers can be reused across batches during sampling while keeping the denoising network frozen. Notably, D-ODE solvers consistently improve the FID of previous ODE solvers, including first-order and high-order methods, significantly reducing the computational time of distillation. Our main contributions can be summarized as follows:

\begin{itemize}
    \item We introduce D-ODE solvers, transferring knowledge from ODE solvers with larger steps to those with smaller steps through a simple formulation.
    \item D-ODE solvers alleviate the need for extensive parameter updates in pre-trained denoising networks, significantly reducing knowledge distillation time.
    \item In quantitative studies, our new sampler outperforms state-of-the-art ODE solvers in terms of FID scores on several image generation benchmarks.
\end{itemize}

 

\section{Background}\label{sec:background}

\noindent\textbf{Forward and reverse diffusion processes.}
The forward process $\{\boldsymbol{x}_t \in \mathbb{R}^D\}_{t \in [0,T]}$ begins with $\boldsymbol{x}_0$ drawn from the data distribution $p_{data}(\boldsymbol{x})$ and evolves to $\boldsymbol{x}_T$ at timestep $T>0$. Given $\boldsymbol{x}_0$, the distribution of $\boldsymbol{x}_t$ can be expressed as:
\begin{equation}
      q_t(\boldsymbol{x}_t|\boldsymbol{x}_0)=\mathcal{N}(\boldsymbol{x}_t|\alpha_t \boldsymbol{x}_0, \sigma_t^2\boldsymbol{I}),
      \label{eq:forward_process}
\end{equation}
where $\alpha_t\in \mathbb{R}$ and $\sigma_t\in \mathbb{R}$ determine the noise schedule of the diffusion models, with the signal-to-noise ratio (SNR) $\alpha_t^2/\sigma_t^2$ strictly decreasing as $t$ progresses~\citep{kingma2021variational}. This ensures that $q_T(\boldsymbol{x}_T)$, the distribution of $\boldsymbol{x}_T$, approximates pure Gaussian noise in practice.

The reverse process of diffusion models is approximated using a denoising network to iteratively remove noise. Starting from $\boldsymbol{x}_T$, the reverse process is defined with the following transition~\citep{ho2020denoising}:
\begin{equation} 
      p_{\theta}(\boldsymbol{x}_{t-1}|\boldsymbol{x}_t) = \mathcal{N}(\boldsymbol{x}_{t-1}|\boldsymbol{\mu}_\theta(\boldsymbol{x}_t,t), \boldsymbol{\Sigma}_\theta(\boldsymbol{x}_t,t)),
\end{equation}
where $\theta$ represents the trainable parameters in the denoising network, and $\boldsymbol{\mu}_\theta(\boldsymbol{x}_t,t)$ and $\boldsymbol{\Sigma}_\theta(\boldsymbol{x}_t,t)$ are the Gaussian mean and variance estimated by the denoising network $\theta$.

\noindent\textbf{SDE and ODE formulation.}
\citet{song2020score} formulate the forward diffusion process using a stochastic differential equation (SDE) to achieve the same transition distribution as \cref{eq:forward_process}. Given $\boldsymbol{x}_0 \sim p_{data}(\boldsymbol{x})$, the forward diffusion process from timestep $0$ to $T$ is newly defined as:
\begin{equation}
      d\boldsymbol{x}_t = f(t) \boldsymbol{x}_t dt + g(t)d  \boldsymbol{w}_t,
\end{equation}
where $ \boldsymbol{w}_t \in \mathbb{R}^D$ is the standard Wiener process, and $f(t)$ and $g(t)$ are functions of $\alpha_t$ and $\sigma_t$. \citet{song2020score} also introduce the reverse-time SDE based on \citet{anderson1982reverse}, which evolves from timestep $T$ to $0$ given $\boldsymbol{x}_T \sim q_T(\boldsymbol{x}_T)$:
\begin{equation}
     d\boldsymbol{x}_t = [f(t) \boldsymbol{x}_t - g^2(t) \nabla_{\boldsymbol{x}} \log q_t( \boldsymbol{x}_t) ]  dt +  g(t)d \bar{\boldsymbol{w}}_t,
\end{equation}
where $\bar{ \boldsymbol{w}}_t$ is the standard Wiener process in reverse time, and $\nabla_{\boldsymbol{x}} \log q_t( \boldsymbol{x}_t)$ is referred to as the score function~\citep{hyvarinen2005estimation}. The randomness introduced by the Wiener process can be omitted to define the diffusion ordinary differential equation (ODE) in the reverse process, which corresponds to solving the SDE on average. Starting from $\boldsymbol{x}_T \sim q_T(\boldsymbol{x}_T)$, probability flow ODE from timestep $T$ to $0$ advances as follows:
\begin{equation}
      d\boldsymbol{x}_t = [f(t)\boldsymbol{x}_t - \frac{1}{2} g^2(t)\nabla_{\boldsymbol{x}} \log q_t(\boldsymbol{x}_t) ]  dt. 
      \label{eq:ODE_reverse}
\end{equation}
The formulation of the probability flow ODE opens up possibilities for using various ODE solvers to expedite diffusion-based sampling processes~\citep{liu2021pseudo, lu2022dpm, zhang2022fast, karras2022elucidating}.

\noindent\textbf{Denoising score matching.}
To solve \cref{eq:ODE_reverse} during sampling, the score function $\nabla_{\boldsymbol{x}} \log q_t( \boldsymbol{x}_t)$ must be estimated. \citet{ho2020denoising} propose estimating the score function using a noise prediction network $\boldsymbol{\epsilon}_{\boldsymbol{\theta}}$ such that $\nabla_{\boldsymbol{x}} \log q_t( \boldsymbol{x}_t)=-\boldsymbol{\epsilon}_{\boldsymbol{\theta}}(\boldsymbol{x}_t, t)/\sigma_t$ with $\boldsymbol{x}_t=\alpha_t \boldsymbol{x} + \sigma_t \boldsymbol{\epsilon}$. The noise prediction network $\boldsymbol{\epsilon}_{\boldsymbol{\theta}}$ is trained using the $L_2$ norm, given samples drawn from $p_{data}$:
\begin{equation}
     \mathbb{E}_{\boldsymbol{x} \sim p_{data}} \mathbb{E}_{\boldsymbol{\epsilon} \sim \mathcal{N}(\boldsymbol{0}, \sigma_t^2 \boldsymbol{I})} || \boldsymbol{\epsilon}_{\boldsymbol{\theta}} (\alpha_t \boldsymbol{x} + \sigma_t \boldsymbol{\epsilon}, t) - \boldsymbol{\epsilon} ||^2.
    \label{eq:L2_noise_estimator}
\end{equation}
Here, Gaussian noise is added to the data $\boldsymbol{x}$ following the noise schedule $(\alpha_t, \sigma_t)$, and the noise prediction network $\boldsymbol{\epsilon}_{\boldsymbol{\theta}}$ predicts the added noise $\boldsymbol{\epsilon}$ from the noisy sample.

\begin{figure*}[t]
\begin{center}
    \includegraphics[width=0.8\linewidth]{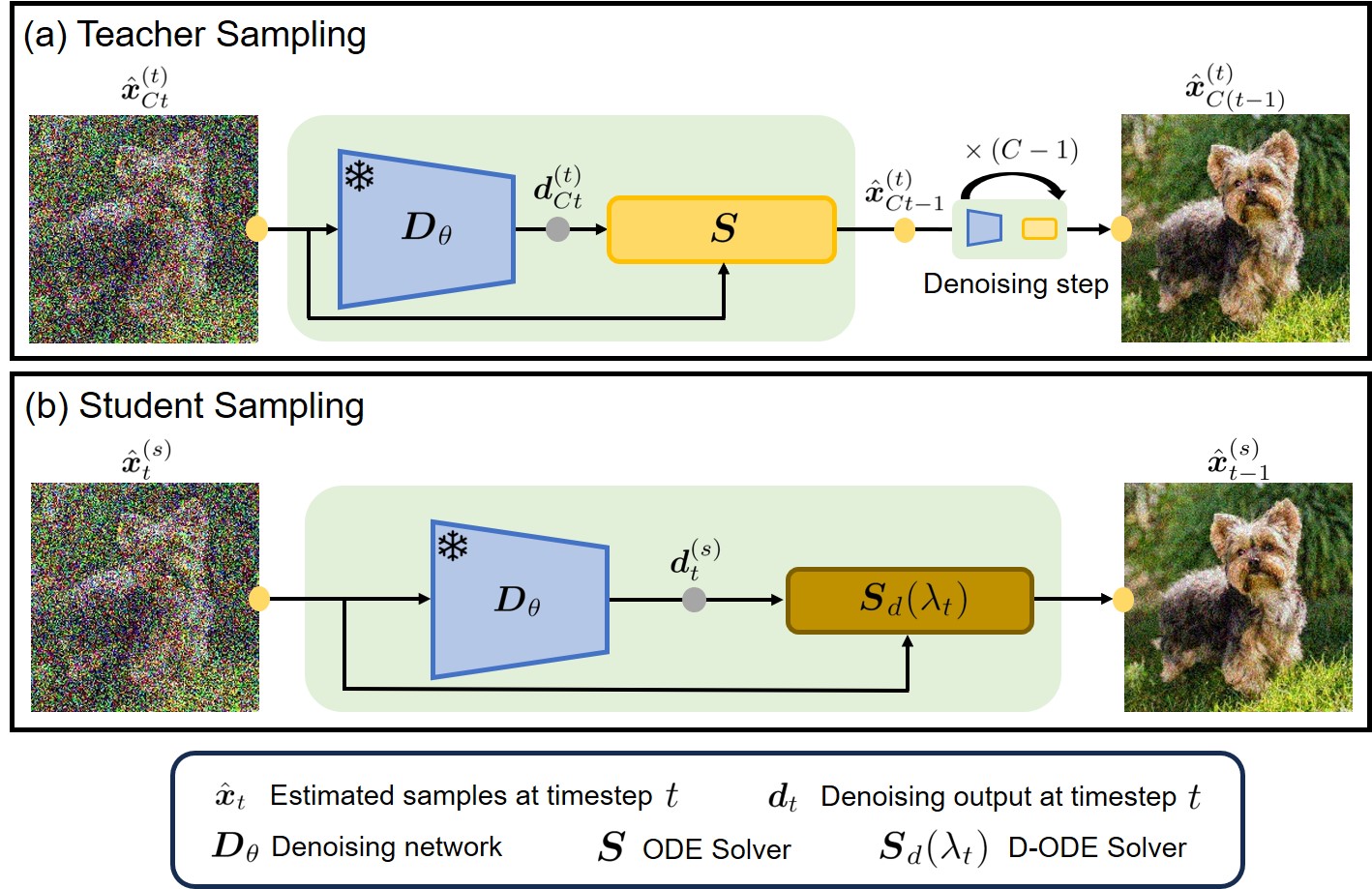}
\end{center}
\caption{\textbf{The overview of D-ODE Solver.} Given an input image at timestep $CT$, teacher sampling performs $C$ denoising steps to obtain the output at time step $C(T-1)$ while student sampling conducts one denoising step from an input at timestep $t$ to an output at timestep $t-1$. Then, $C$ steps of the teacher sampling are distilled into a single step of the student sampling by optimizing $\lambda_t$ within the D-ODE solver. Note that the denoising network remains frozen for both teacher and student sampling.} \label{fig:distillation_explanation}
\vspace{-0.4cm}
\end{figure*}


Alternatively, the score function can be represented using a data prediction network $\boldsymbol{x}_{\boldsymbol{\theta}}$ instead of $\boldsymbol{\epsilon}_{\boldsymbol{\theta}}$ with $\nabla_{\boldsymbol{x}} \log q_t( \boldsymbol{x}_t) = (\boldsymbol{x}_{\boldsymbol{\theta}}(\boldsymbol{x}_t, t) - \boldsymbol{x}_t)/\sigma_t^2$. The data prediction network $\boldsymbol{x}_{\boldsymbol{\theta}}$ is trained with following $L_2$ norm:
\begin{equation}
     \mathbb{E}_{\boldsymbol{x} \sim p_{data}} \mathbb{E}_{\boldsymbol{\epsilon} \sim \mathcal{N}(\boldsymbol{0}, \sigma_t^2 \boldsymbol{I})} || \boldsymbol{x}_{\boldsymbol{\theta}} (\alpha_t \boldsymbol{x} + \sigma_t \boldsymbol{\epsilon}, t) - \boldsymbol{x} ||^2.
\end{equation}
It is worth noting that estimating the original data $\boldsymbol{x}$ is theoretically equivalent to learning to predict the noise $\boldsymbol{\epsilon}$~\citep{ho2020denoising, luo2022understanding}. While some works argue that predicting the noise empirically results in higher quality samples~\citep{ho2020denoising, saharia2022photorealistic}, \citet{karras2022elucidating} recently achieved state-of-the-art performance using the data prediction network. In this work, we conduct comprehensive experiments with both noise and data prediction networks. For the rest of the paper, we write $\boldsymbol{D}_\theta$ to represent the denoising network of the diffusion models which can be either noise or data prediction networks.

\section{The Proposed Method}

Our study aims to bridge the gap between learning-based and learning-free sampling, leveraging the advantages of both approaches. We capitalize on the sampling dynamics of ODE solvers while enhancing sample quality through a straightforward and efficient knowledge distillation. This section begins with a fundamental observation of the high correlation among the outputs of the denoising network (\ie, denoising output), motivating the formulation of D-ODE solvers. We then delve into the details of transferring knowledge from ODE solvers to D-ODE solvers.

\subsection{Correlation between Denoising Outputs}
\begin{figure}[h]
\begin{center}
\includegraphics[width=1.0\linewidth]{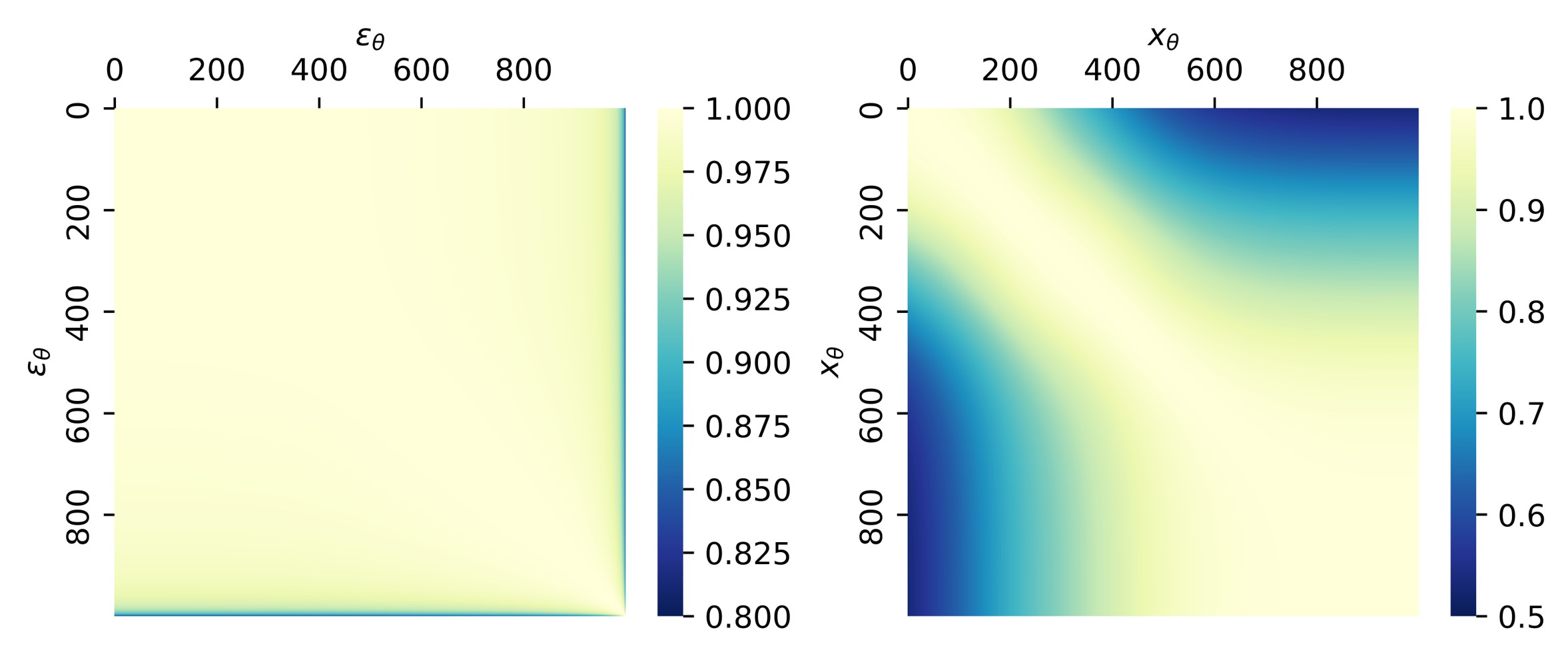} 
\end{center}
\caption{\textbf{Correlation between denoising outputs.} Heatmaps are drawn by cosine similarity among denoising outputs with 1000-step DDIM on CIFAR-10. Noise prediction model (left) and data prediction model (right).} \label{fig:cossim_cifar10}
\vspace{-0.5cm}
\end{figure}

ODE solvers typically enhance the sampling process by exploiting the output history of the denoising network, enabling the omission of many intermediate steps. Therefore, understanding the relationship between denoising outputs is crucial when developing D-ODE Solvers. Our objective is to create novel ODE solvers that harness the benefits of sampling dynamics while keeping the degrees of optimization freedom to a minimum.

\cref{fig:cossim_cifar10} presents heatmaps based on cosine similarity calculations between all denoising outputs from a 1000-step DDIM~\citep{song2020denoising} run. We observe that the denoising outputs from neighboring timesteps exhibit high correlations in both noise and data prediction models, with cosine similarities close to one. This observation suggests that denoising outputs contain redundant and duplicated information, allowing us to skip the evaluation of denoising networks for most timesteps. For example, the history of denoising outputs can be combined to better represent the next output, effectively reducing the number of steps required for accurate sampling. This idea is implemented in most ODE solvers, which are formulated based on the theoretical principles of solving differential equations~\cite{watson2021learning, liu2021pseudo, karras2022elucidating, lu2022dpm, zhang2022fast}. 


\subsection{Formulation of D-ODE Solver}\label{subsec:formulation_d_ode_solver}
As illustrated in \cref{fig:distillation_explanation}, each denoising step in diffusion models typically involves two components: (1) a denoising network $\boldsymbol{D}_{\boldsymbol{\theta}}$ and (2) an ODE solver $\boldsymbol{S}$. Given an estimated noisy sample $\hat{\boldsymbol{x}}_{t}$ at timestep $t$, the denoising network $\boldsymbol{D}_{\boldsymbol{\theta}}$ produces a denoising output $\boldsymbol{d}_t=\boldsymbol{D}_{\boldsymbol{\theta}}(\hat{\boldsymbol{x}}_{t}, t)$, and the ODE solver subsequently generates the next sample $\hat{\boldsymbol{x}}_{t-1}=\boldsymbol{S}(\boldsymbol{d}_t, \hat{\boldsymbol{x}}_{t})$, utilizing the denoising output and the noisy sample at timestep $t$. While high-order ODE solvers also utilize the history of denoising outputs $\{\boldsymbol{d}_k\}_{k=t}^T$, we omit this notation here for simplicity. This procedure is iterated until the diffusion models reach the estimated original sample $\hat{\boldsymbol{x}}_{0}$.

We now introduce a D-ODE solver with a straightforward parameterization to distill knowledge from ODE solvers. We begin by outlining a fundamental method to estimate the new denoising output $\boldsymbol{O}_t$ at timestep $t$ as a linear combination of current and previous denoising outputs $\{\boldsymbol{d}_k\}_{k=t}^T$:
\begin{equation}
    \boldsymbol{O}_t = \sum_{k=t}^{T} \lambda_k \boldsymbol{d}_k,
    \label{eq:linear_combination}
\end{equation}
where $\lambda_k \in \mathbb{R}$ is a weight parameter for each denoising output $\boldsymbol{d}_k$. With carefully chosen $\lambda_k$, we anticipate that the new denoising output can better approximate the target score function of ODE in \cref{eq:ODE_reverse}, leading to improved sample quality. Some high-order ODE solvers~\citep{liu2021pseudo, lu2022dpm, zhang2022fast} adopt similar formulations to \cref{eq:linear_combination} with mathematically determined weight parameters $\{\lambda_k\}_{k=t}^T$.

One challenge within \cref{eq:linear_combination} is that the value of the new denoising output $\boldsymbol{O}_t$ can be unstable and volatile depending on the values of weights $\{\lambda_k\}_{k=t}^T$. This instability is less likely to occur with numerically computed weights in ODE solvers, but convergence is not guaranteed when the weights are optimized through knowledge distillation. To generate high-quality samples, the sampling process must follow the true ODE trajectory on which the diffusion models are trained~\citep{liu2021pseudo, song2023consistency}. In other words, the denoising network might not produce reliable outputs for samples outside the target manifold of data~\citep{xiao2021tackling, ning2023input, li2023alleviating}. 

To avoid this, \cref{eq:linear_combination} should be constrained so that it adheres to the original ODE trajectory. Thus, the new denoising output $\boldsymbol{O}_t$ can be defined as follows:
\begin{align}
        \boldsymbol{O}_t &= \boldsymbol{d}_t + \sum_{k=t+1}^{T} \lambda_{k-1} (\boldsymbol{d}_t - \boldsymbol{d}_k)  \label{eq:first_formulation_d_ode}  \\
        &\approx \boldsymbol{d}_t + \lambda_{t} (\boldsymbol{d}_t - \boldsymbol{d}_{t+1}).
        \label{eq:final_formulation_d_ode}
\end{align}

Furthermore, we empirically find that using the denoising output from the previous timestep is sufficient for distilling knowledge from the teacher sampling (see Supplementary Material). As a result, we obtain \cref{eq:final_formulation_d_ode} for D-ODE solvers. It is worth noting that the mean of the new denoising output $\boldsymbol{O}_t$ approximates that of the original denoising output since the mean with respect to sample $\boldsymbol{x}$ in sufficiently large batch does not change significantly between timesteps $t$ and $t+1$ (\eg, $\mathbb{E}_{\boldsymbol{x} \sim p_{data}}[\boldsymbol{O}_t] \approx \mathbb{E}_{\boldsymbol{x} \sim p_{data}}[\boldsymbol{d}_t]$). This is a key feature of D-ODE solvers, as we aim to remain on the original sampling trajectory of ODE.

In case of DDIM~\citep{song2020denoising}, one can simply substitute $\boldsymbol{d}_t$ with $\boldsymbol{O}_t$ to construct D-DDIM:
\begin{align}
    \text{DDIM:} \hspace{0.5em} &\hat{\boldsymbol{x}}_{t-1} = \alpha_{t-1}  \left(\frac{\hat{\boldsymbol{x}}_t-\sigma_t \boldsymbol{d}_t}{\alpha_t} \right) +  \sigma_{t-1}\boldsymbol{d}_t,\\
    \text{D-DDIM:} \hspace{0.5em} &\hat{\boldsymbol{x}}_{t-1} = \alpha_{t-1}  \left(\frac{\hat{\boldsymbol{x}}_t-\sigma_t \boldsymbol{O}_t}{\alpha_t} \right) +  \sigma_{t-1}\boldsymbol{O}_t.  
\end{align}
where ($\alpha_t$, $\sigma_t$) represents a predefined noise schedule. $\lambda_{t}$ is optimized later via knowledge distillation.

\noindent\textbf{Comparison with high-order ODE solvers.} High-order methods for sampling utilize the history of denoising outputs. As these methods better approximate the target score function of ODE compared to the first-order method (\eg, DDIM), we apply \cref{eq:final_formulation_d_ode} on top of their approximation to build D-ODE solvers. In other words, $\boldsymbol{d}_t$ in \cref{eq:final_formulation_d_ode} is replaced by the high-order approximation of each method. In this way, we can involve more timesteps to obtain $\boldsymbol{O}_t$ while overcoming the bottleneck of ODE solvers with an extra parameter $\lambda_t$ adapted to each dataset. Unlike high-order ODE solvers, D-ODE solvers are equipped with the parameter $\lambda_t$, optimized for a specific dataset through knowledge distillation, to further reduce the fitting error of the score function. Supplementary Material includes the specific applications of D-ODE solvers and different formulations of D-ODE solvers.


\subsection{Knowledge Distillation of D-ODE Solver}
In \cref{fig:distillation_explanation}, the teacher sampling process initiates with the noisy sample $\hat{\boldsymbol{x}}^{(t)}_{CT}$ at timestep $Ct$ and undergoes $C$ denoising steps to generate a sample $\hat{\boldsymbol{x}}^{(t)}_{C(T-1)}$ at timestep $C(t-1)$. Simultaneously, the student sampling process commences with a noisy sample $\hat{\boldsymbol{x}}^{(s)}_{t}$ at timestep $t$ and obtains a sample $\hat{\boldsymbol{x}}^{(s)}_{t-1}$ at timestep $t-1$ after one denoising step. To optimize $\lambda_t$ in the D-ODE solver $\boldsymbol{S}_d$, the teacher sampling is initially conducted for one batch, saving intermediate samples $\{\hat{\boldsymbol{x}}^{(t)}_{k}\}_{k=C(t-1)}^{Ct}$ as targets. The student sampling is also performed, obtaining intermediate samples $\{\hat{\boldsymbol{x}}^{(s)}_{k}\}_{k=t-1}^t$ as predictions. Subsequently, $\lambda_t^*$ is determined by minimizing the difference between the targets and predictions on batch $B$ as follows:

\begin{align}
    \lambda_t^* &= \argmin_{\lambda_t} \mathbb{E}_{\boldsymbol{x} \in B} ||\hat{\boldsymbol{x}}^{(t)}_{C(t-1)}-S_d(\boldsymbol{d}_t^{(s)},\hat{\boldsymbol{x}}^{(s)}_{t}; \lambda_t)||^2 \\
                &= \argmin_{\lambda_t} \mathbb{E}_{\boldsymbol{x} \in B}||\hat{\boldsymbol{x}}^{(t)}_{C(t-1)}-\hat{\boldsymbol{x}}^{(s)}_{t-1}||^2, \label{eq:argmin_lambda}
\end{align} 
where $\boldsymbol{d}_t^{(s)} = \boldsymbol{D}_{\boldsymbol{\theta}}(\hat{\boldsymbol{x}}^{(s)}_{t},t)$ holds.
The above equation is solved for every timestep $t$ of the student sampling, yielding a set of optimal $\lambda_t$ values (\eg, $\lambda^* =\{\lambda_0^*, \lambda_1^*, ..., \lambda_{T-1}^* \}$). Notably, $\lambda^*$ is estimated using only one batch of samples, a process that typically takes just a few CPU minutes, and can be reused for other batches later. 

\cref{alg:sampling} outlines the overall sampling procedure of the D-ODE solver. When generating $N$ samples, it is normal to divide $N$ into $M$ batches and sequentially execute the sampling process for each batch $B$, which contains $|B| = N/M$ samples (Line 3). For the first batch, teacher sampling is conducted with denoising network $\boldsymbol{D}_{\boldsymbol{\theta}}$ and ODE solver $\boldsymbol{S}$ for $CT$ steps to obtain intermediate outputs, which will serve as target samples (Line 7). Subsequently, student sampling takes place for $T$ steps with D-ODE Solver $\boldsymbol{S}_d(\lambda)$ (Line 8). At this point, $\lambda^*$ is estimated and saved for each timestep by solving \cref{eq:argmin_lambda} (Line 9). Starting from the second batch onwards, sampling can proceed using the frozen denoising network $\boldsymbol{D}_{\boldsymbol{\theta}}$ and D-ODE solver $\boldsymbol{S}_d(\lambda^*)$ (Line 9). It is important to note that the student's samples can be generated in just $T$ steps, which exhibits similar quality to the teacher's samples generated over $CT$ steps. 

\begin{algorithm}
  \caption{Sampling with D-ODE solver}\label{alg:sampling}  
  \begin{algorithmic}[1]
  \State Pre-trained denoising network $\boldsymbol{D}_{\boldsymbol{\theta}}$
  \State ODE solver $\boldsymbol{S}$, D-ODE solver $\boldsymbol{S}_d (\lambda)$
  \State Number of batches $M$ with size $|B|$
  \State Student sampling steps $T$, Teacher sampling steps $CT$
  \For{$m = 1,... , M$}
    \If {$m = 1$}
    \State $\{\hat{\boldsymbol{x}}^{(t)}_{k}\}_{k=0}^{CT}$ = \textit{Teacher-Sampling}($\boldsymbol{D}_\theta$, $\boldsymbol{S}$, $CT$)   
    \State $\{\hat{\boldsymbol{x}}^{(s)}_{k}\}_{k=0}^{T}$ = \textit{Student-Sampling}($\boldsymbol{D}_\theta$, $\boldsymbol{S}_d(\lambda)$, $T$) 
    \State Estimate $\lambda^* =\{\lambda_1^*, \lambda_2^*, ..., \lambda_T^* \}$ with \cref{eq:argmin_lambda}
    \EndIf
  \State $\{\hat{\boldsymbol{x}}^{(s)}_{k}\}_{k=0}^{T}$ = \textit{Student-Sampling}($\boldsymbol{D}_\theta$, $\boldsymbol{S}_d(\lambda^*)$, $T$)
  \State Save sample $\hat{\boldsymbol{x}}^{(s)}_{0}$ 
  \EndFor 
  \end{algorithmic}
\end{algorithm}


\section{Experiments}
In this section, we present a comprehensive evaluation of D-ODE solvers in comparison to ODE solvers on diverse image generation benchmarks at various resolutions, including CIFAR-10 ($32\times32$), CelebA ($64\times64$), ImageNet ($64\times64$ and $128\times128$), FFHQ ($64\times64$), and LSUN bedroom ($256\times256$). Our experiments cover both noise and data prediction models, each involving distinct sets of ODE solvers. The Fréchet Inception Distance (FID)~\citep{heusel2017gans} is employed as the evaluation metric, measured with 50K generated samples across various numbers of denoising function evaluations (NFE), following the protocol of \citet{lu2022dpm}. The reported FID scores are averaged over three independent experiment runs with different random seeds.

For the distillation of ODE solvers, we set the scale parameter to $C=10$ and use a batch size of $|B|=100$, except for the LSUN bedroom dataset, where a batch size of 25 is employed due to GPU memory constraints. It is important to note that, unless explicitly specified, DDIM serves as the primary teacher sampling method to guide the student sampling. This choice is made considering that certain ODE solvers employ multi-step approaches during sampling, making it challenging to set their intermediate outputs as targets for distillation. In contrast, DDIM generates a single intermediate output per denoising step, simplifying the establishment of matching pairs between DDIM targets and student predictions. Refer to the Supplementary Material for detailed applications of D-ODE solvers and ablation studies on the scale $C$ and batch size $|B|$.

\begin{figure*}[t]
	\centering
	\begin{subfigure}[t]{0.3\linewidth}
		\includegraphics[width=\linewidth]{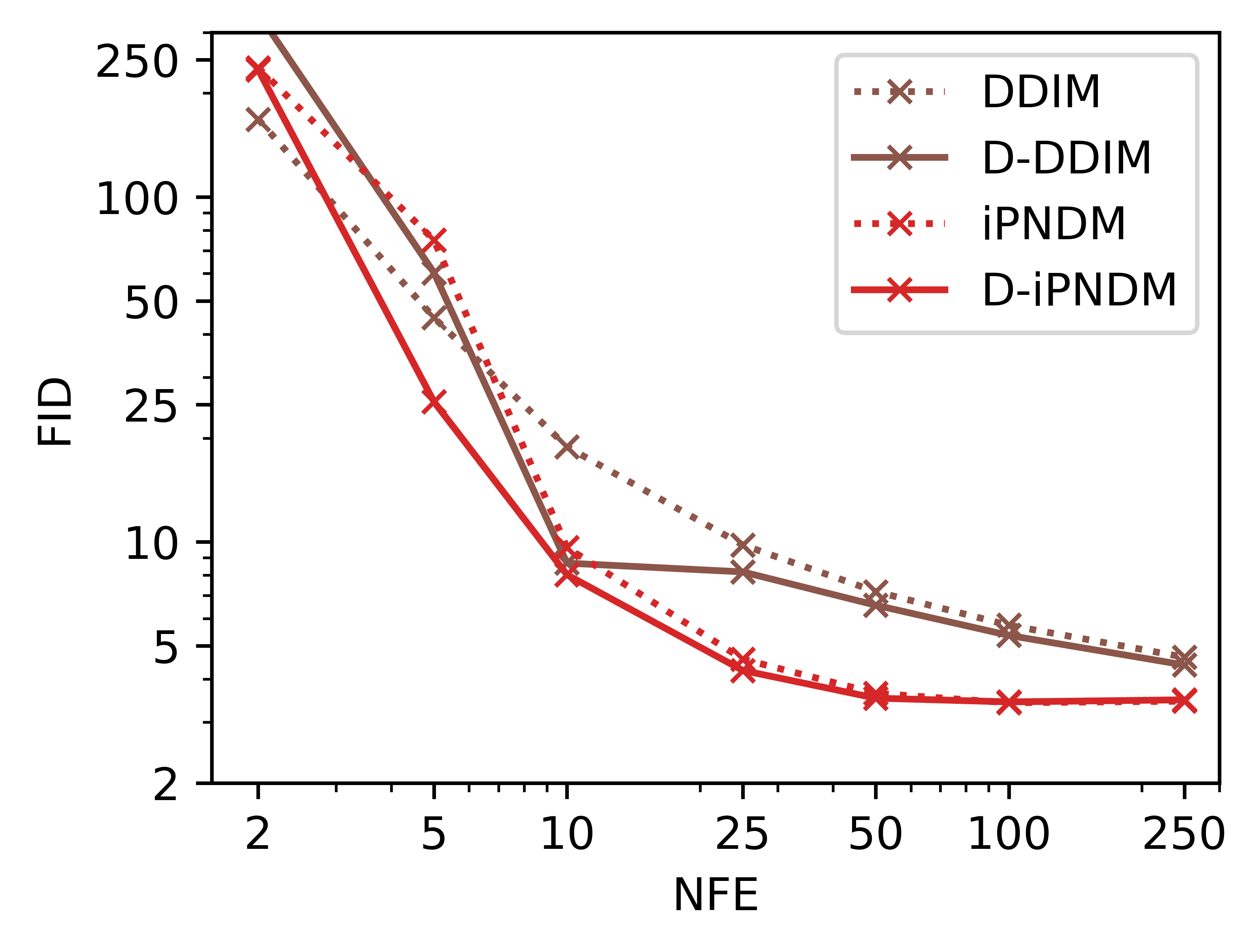}
		\caption{CIFAR-10 $(32\times32)$}
		\label{fig:noise_subfigA}
	\end{subfigure}
	\begin{subfigure}[t]{0.3\linewidth}
		\includegraphics[width=\linewidth]{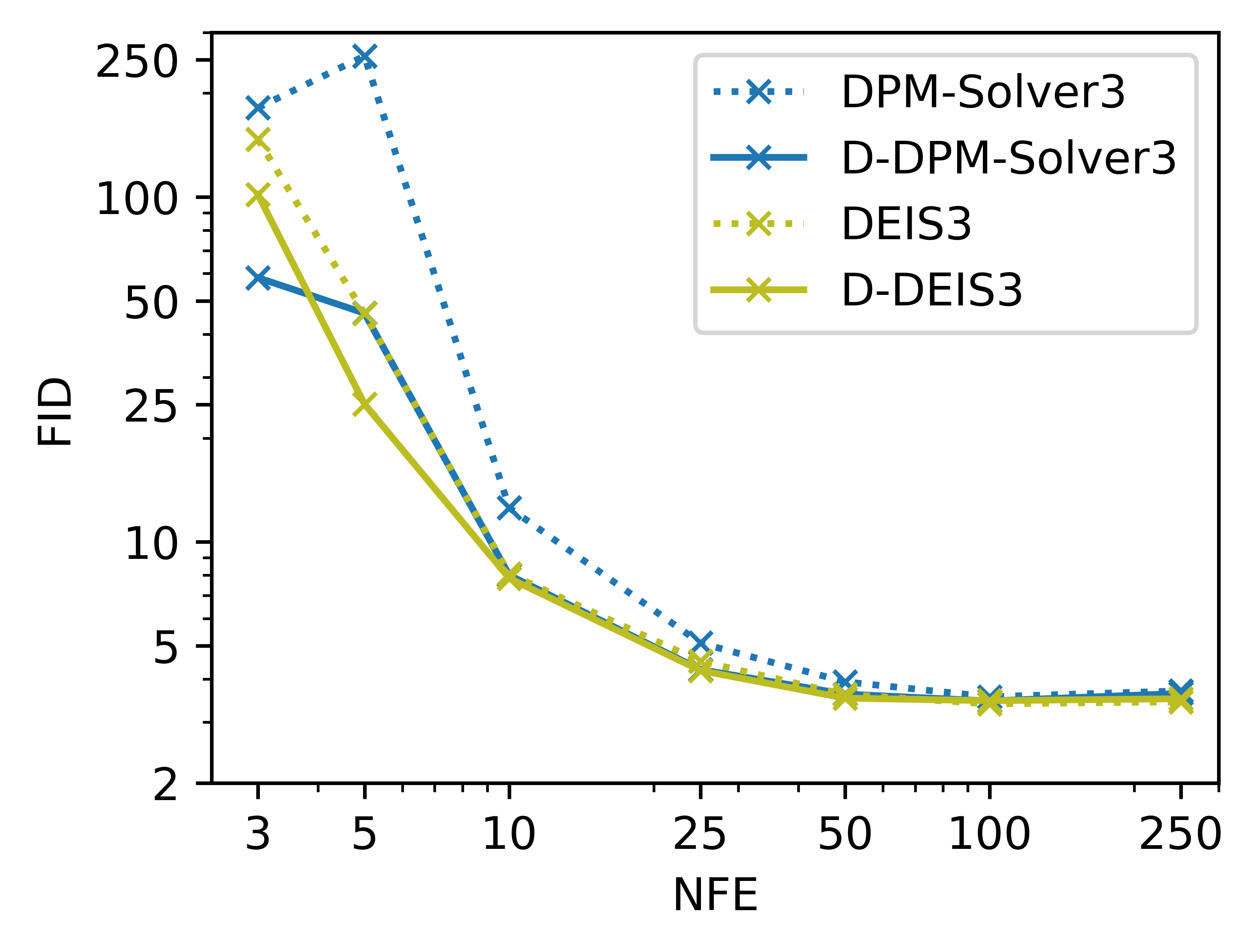}
		\caption{CIFAR-10 $(32\times32)$}
		\label{fig:noise_subfigB}
	\end{subfigure}
	\begin{subfigure}[t]{0.3\linewidth}
            \includegraphics[width=\linewidth]{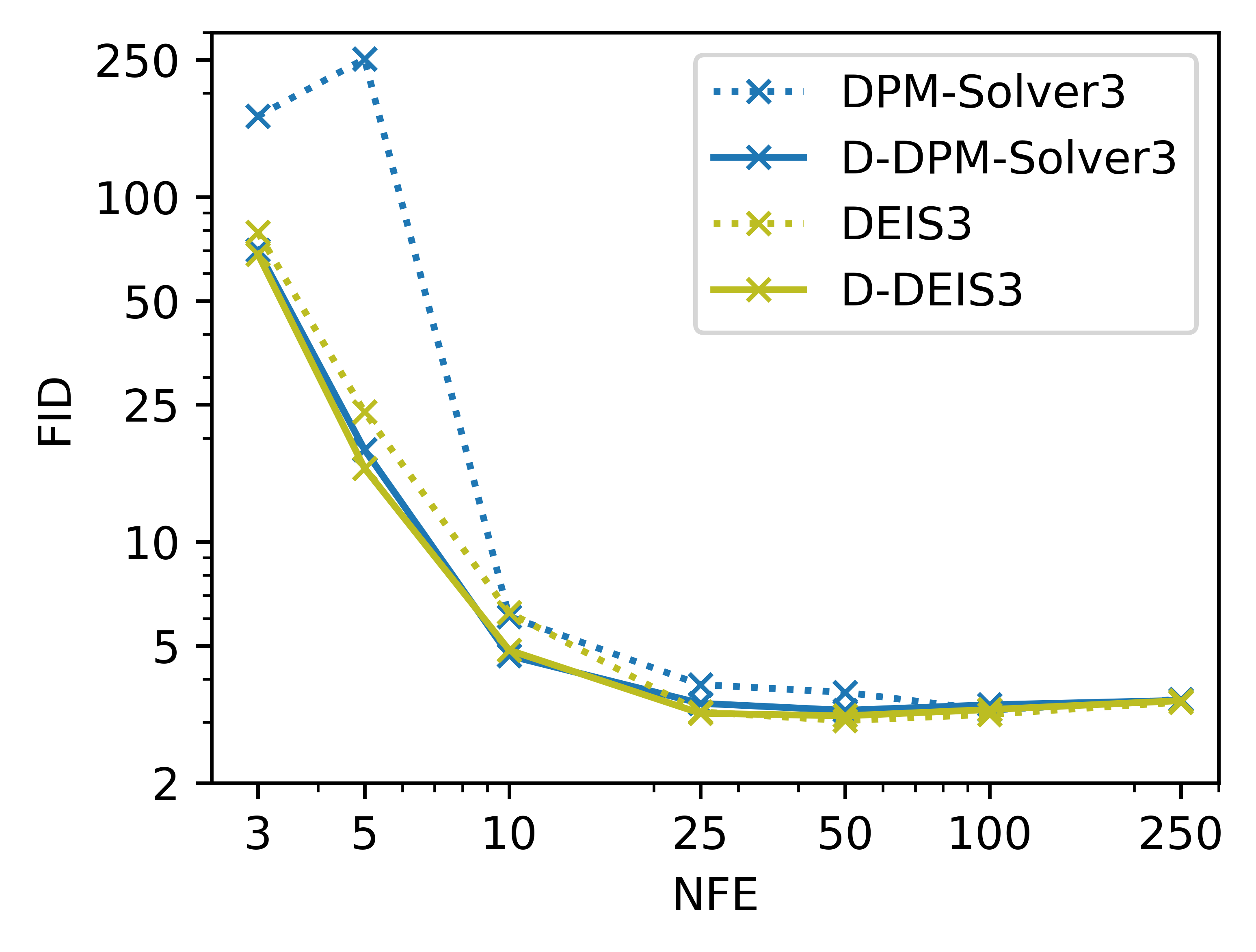}
            \caption{ImageNet $(128\times128)$}
            \label{fig:noise_subfigC}
        \end{subfigure} \\
        
	\begin{subfigure}[t]{0.3\linewidth}
		\includegraphics[width=\linewidth]{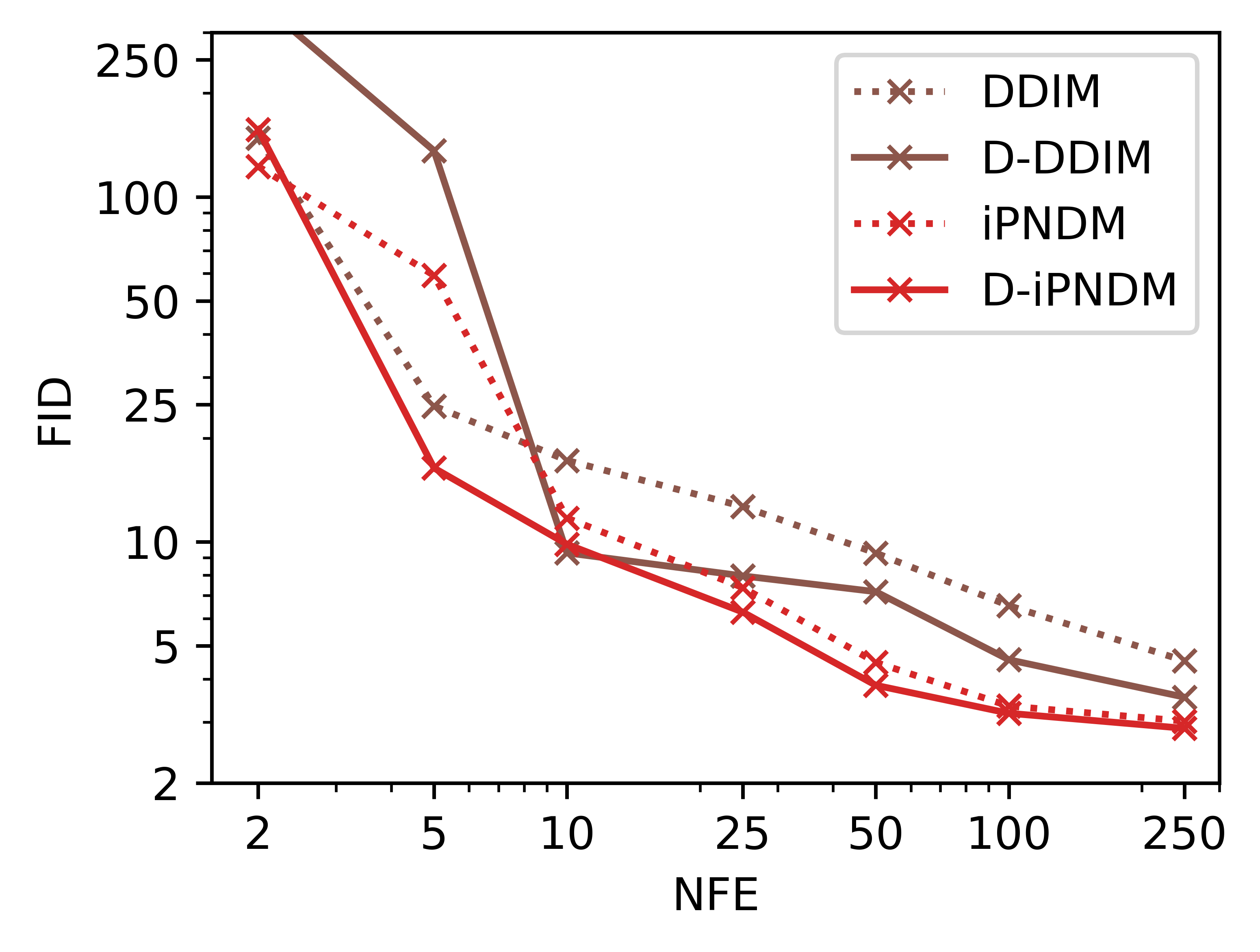}
		\caption{CelebA $(64\times64)$}
		\label{fig:noise_subfigD}
	\end{subfigure}
	\begin{subfigure}[t]{0.3\linewidth}
		\includegraphics[width=\linewidth]{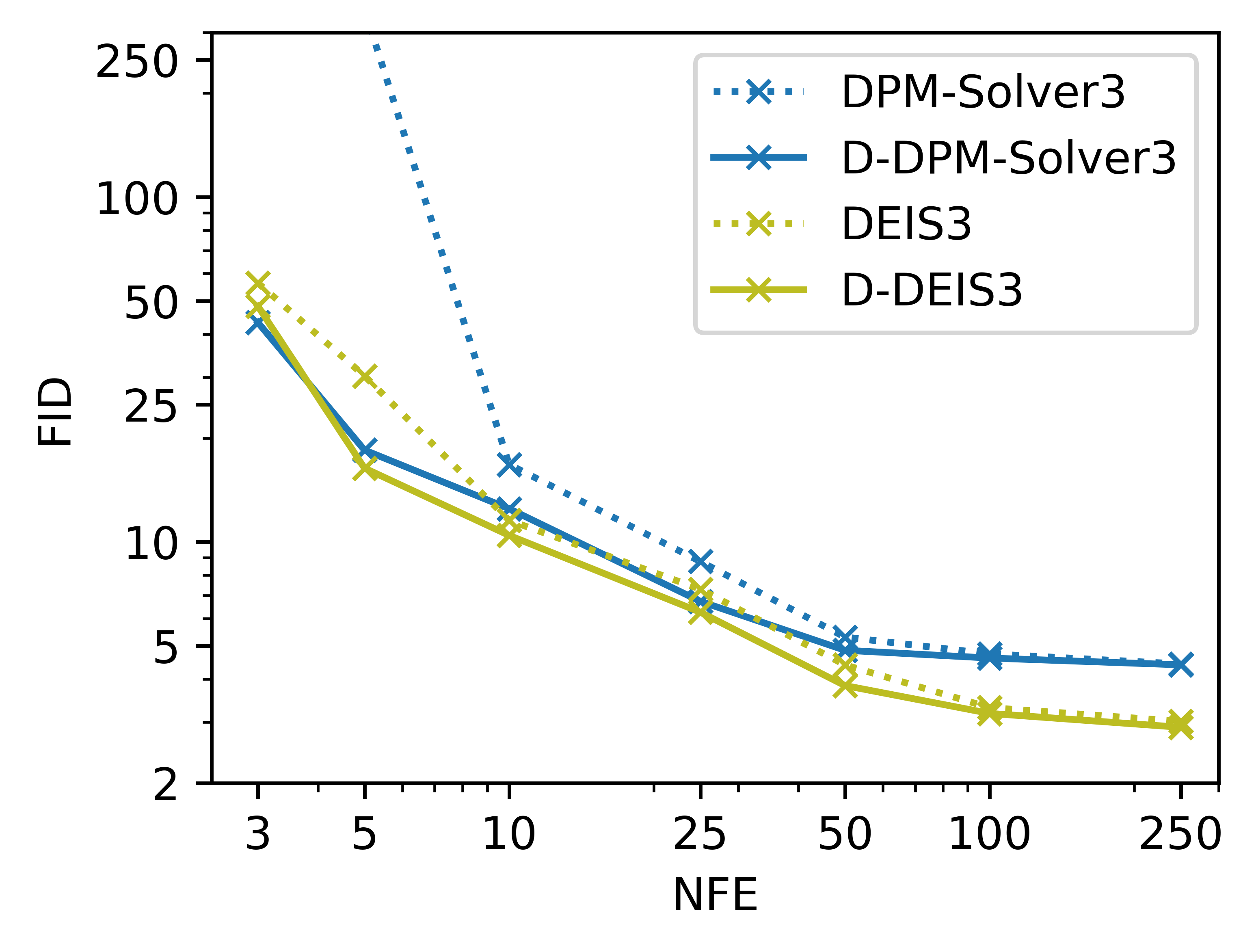}
		\caption{CelebA $(64\times64)$}
		\label{fig:noise_subfigE}
	\end{subfigure}
	\begin{subfigure}[t]{0.3\linewidth}
            \includegraphics[width=\linewidth]{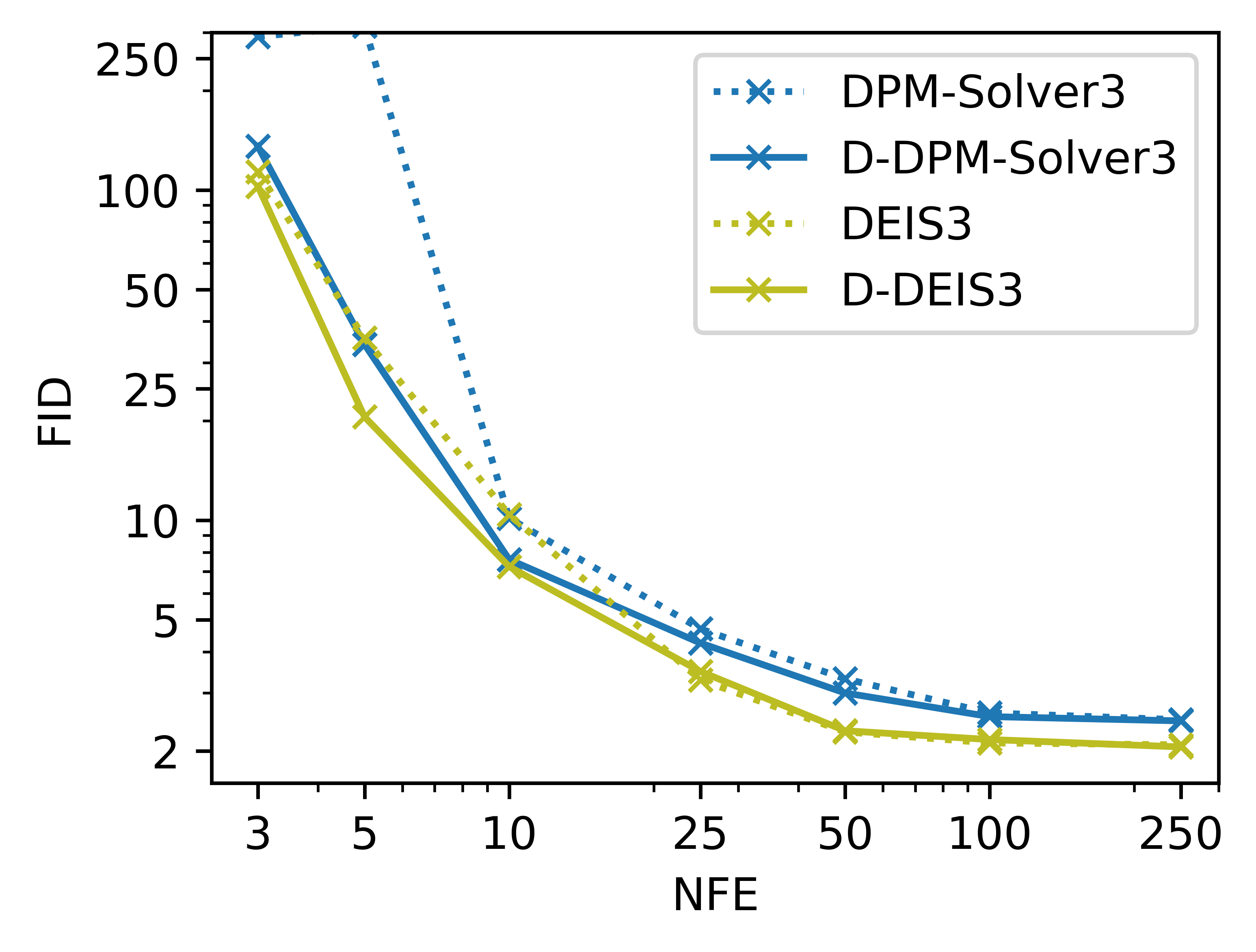}
            \caption{LSUN bedroom $(256\times256)$}
            \label{fig:noise_subfigF}
        \end{subfigure}
	\caption{\textbf{Results on the noise prediction models.} Image quality measured by FID $\downarrow$ with NFE $\in \{2, 5, 10, 25, 50, 100, 250\}$. For DPM-Solver3 and DEIS3, we use 3 NFE instead of 2 NFE as the third-order method requires at least three denoising outputs. Dotted lines denote ODE solvers while straight lines represent the applications of the D-ODE solver to them.}
	\label{fig:noise_subfigures}
 \vspace{-0.4cm}
\end{figure*}

\begin{figure*}
	\centering
	\begin{subfigure}[t]{0.3\linewidth}
		\includegraphics[width=\linewidth]{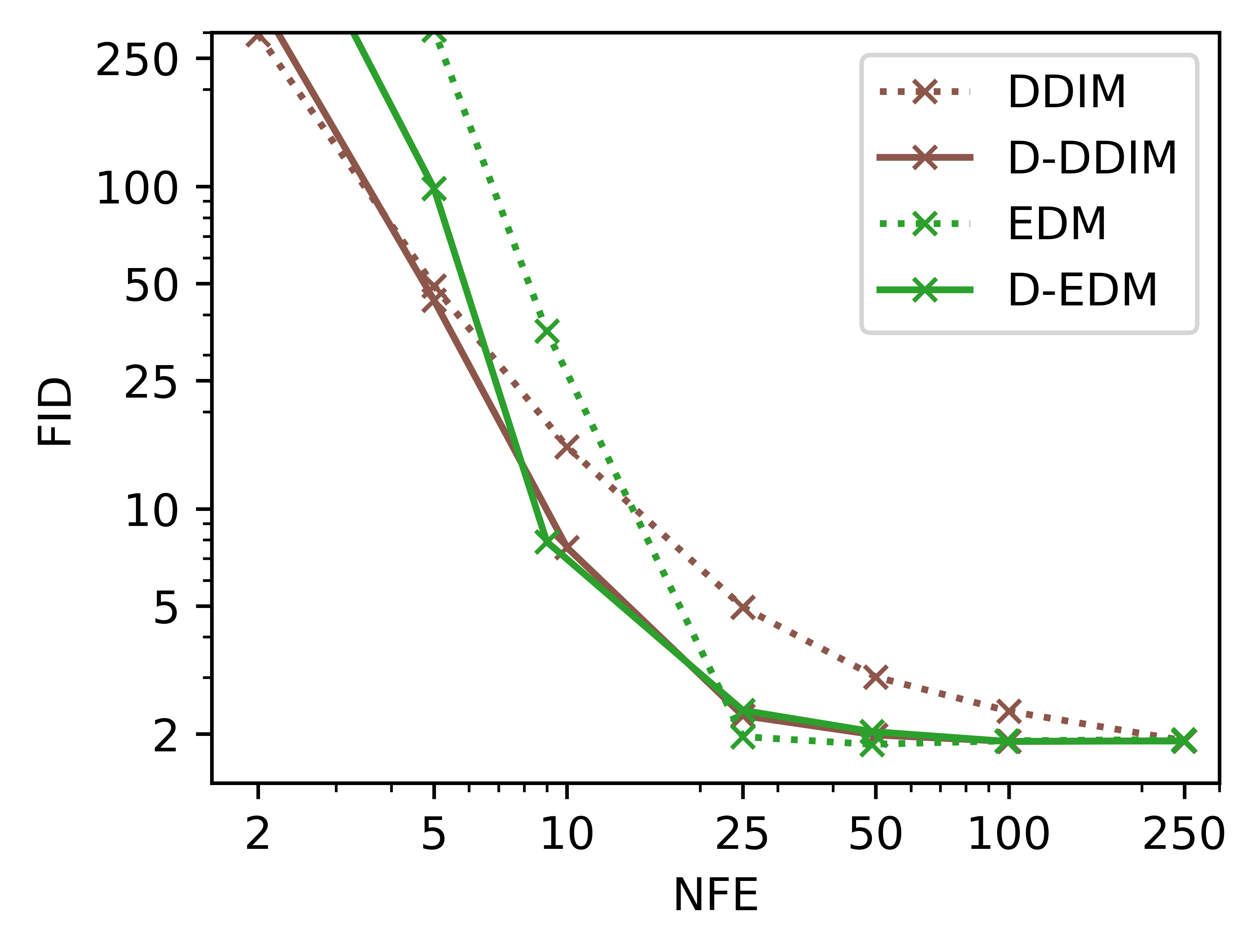}
		\caption{CIFAR-10 $(32\times32)$}
		\label{fig:sample_subfigA}
	\end{subfigure}
	\begin{subfigure}[t]{0.3\linewidth}
		\includegraphics[width=\linewidth]{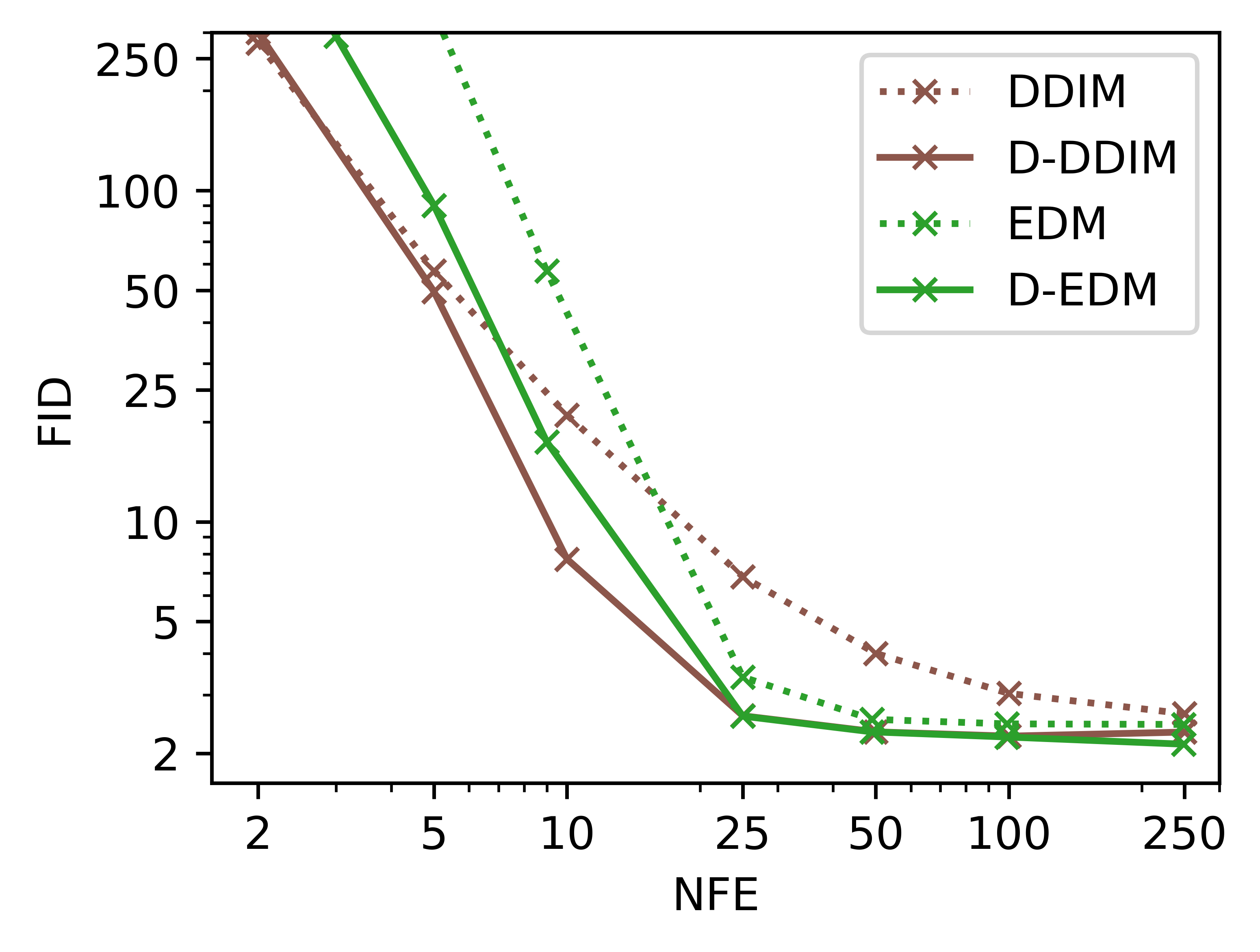}
		\caption{FFHQ $(64\times64)$}
		\label{fig:sample_subfigB}
	\end{subfigure}
	\begin{subfigure}[t]{0.3\linewidth}
	        \includegraphics[width=\linewidth]{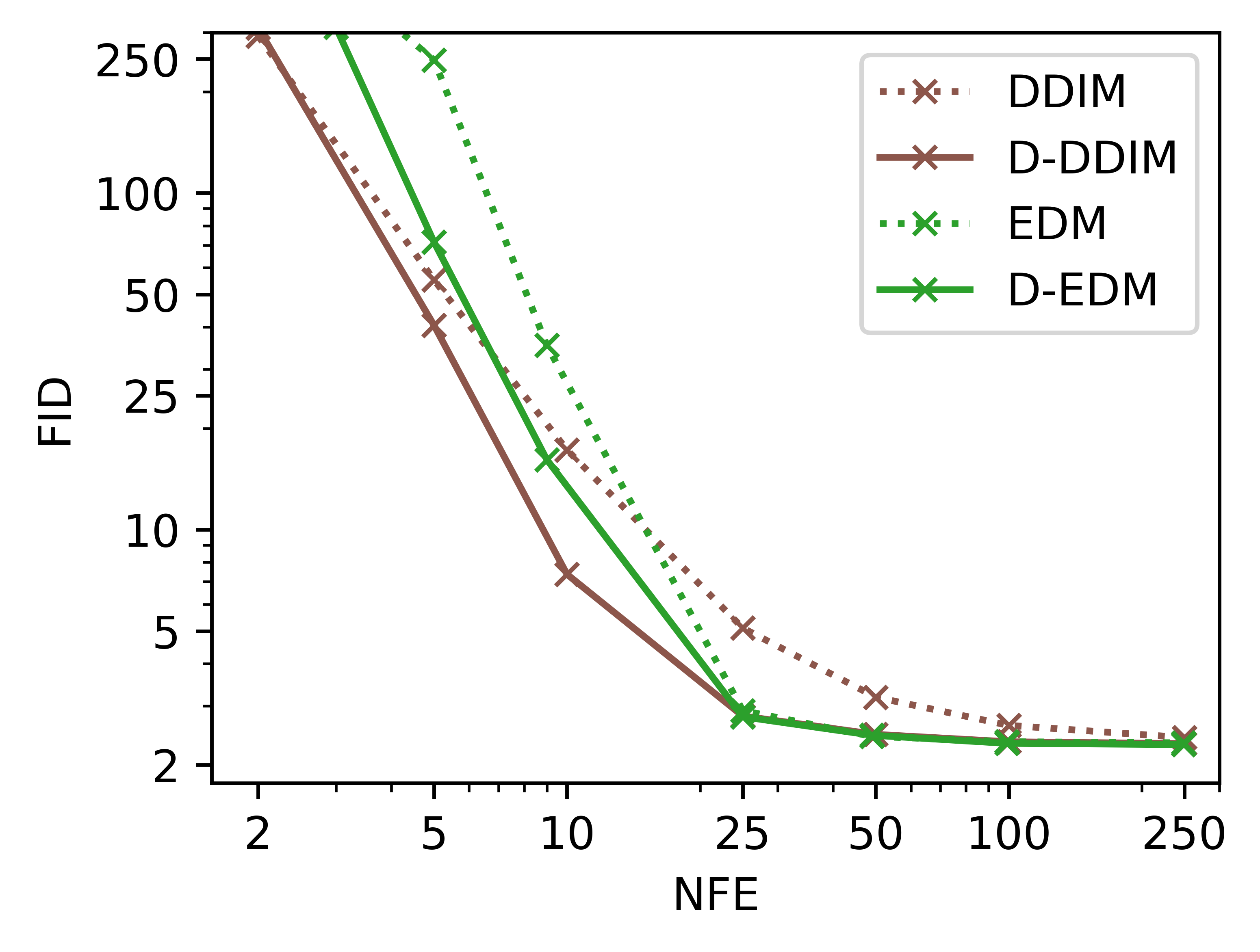}
	        \caption{ImageNet $(64\times64)$}
	        \label{fig:sample_subfigC}
         \end{subfigure}
	\caption{\textbf{Results on the data prediction models.} Image quality measured by FID $\downarrow$ with various NFE values (DDIM: \{2, 5, 10, 25, 50, 100, 250\} and EDM: \{3, 5, 9, 25, 49, 99, 249\}). Dotted lines denote ODE solvers and straight lines represent the applications of the D-ODE solver to them.}
	\label{fig:sample_subfigures}
 \vspace{-0.4cm}
\end{figure*}

\subsection{Noise Prediction Model}

We apply D-ODE solvers to discrete-time ODE solvers employed in the noise prediction model, including DDIM~\citep{song2020denoising}, iPNDM~\citep{zhang2022fast}, DPM-Solver~\citep{lu2022dpm}, and DEIS~\citep{zhang2022fast}. For DPM-Solver and DEIS, we selected third-order methods. While these ODE solvers were primarily evaluated with NFE greater than 10, we also conduct experiments with extremely small NFE, such as 2 or 3, to assess the performance of D-ODE solvers during the initial stages of the sampling process.

\cref{fig:noise_subfigures} illustrates that D-ODE solvers outperform ODE solvers, achieving lower FID in most NFEs. In \cref{fig:noise_subfigA} and \cref{fig:noise_subfigD}, D-DDIM outperforms DDIM when NFE exceeds 5, gradually converging to FID score similar to that of DDIM as NFE increases. It is important to note that DDIM with small NFE (2 or 5) lacks the capability to produce meaningful images, which is also reflected in the performance of D-DDIM. iPNDM, a high-order method that utilizes previous denoising outputs, consistently exhibits improvements with the D-ODE solver formulation, except at 2 NFE. This improvement is particularly notable for high-order methods like DPM-Solver3 and DEIS3. Specifically, D-DPM-Solver3 effectively alleviates the instability associated with multi-step approaches at extremely small NFE values, surpassing the performance of DPM-Solver3 by a significant margin. While DEIS3 already provides a precise representation of the current denoising output through high-order approximation, \cref{fig:noise_subfigures} illustrates that D-DEIS3 can further enhance the approximation with parameter $\lambda$ optimized for each dataset through knowledge distillation. In Supplementary Matrial, we also show that applying D-ODE solvers is effective for DPM-Solver++~\cite{lu2022dpm++}.

\begin{figure*}[h] 
\begin{center}
    \includegraphics[width=1.0\linewidth]{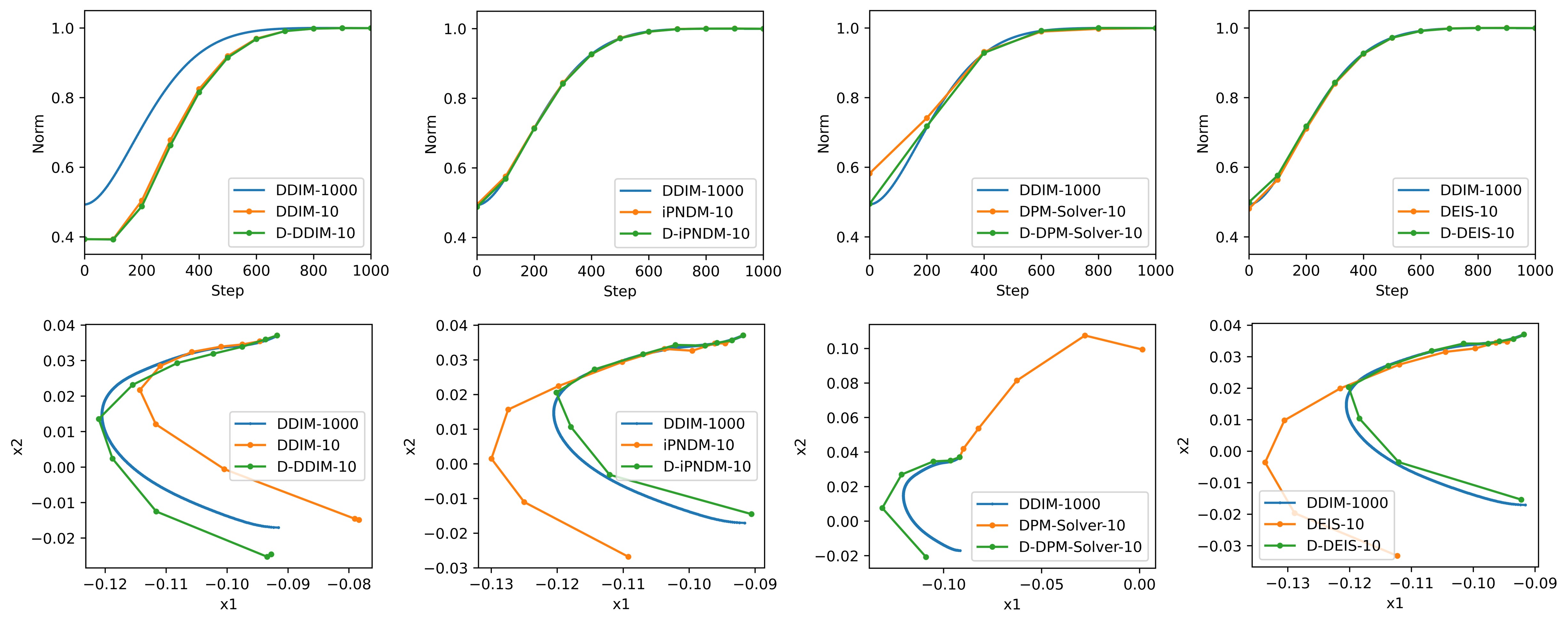}
\end{center}
\vspace{-1em}
\caption{\textbf{Analysis on local and global characteristics.} The top row illustrates the change of norm comparing ODE and D-ODE solvers. The bottom row presents the update path of two randomly selected pixels in the images. The result of \textcolor{blue}{1000-step DDIM} is drawn as the target trajectory and a 10-step sampler is conducted for \textcolor{orange}{ODE solvers} and \textcolor{ForestGreen}{D-ODE solvers}. The figures are generated from 1000 samples using a noise prediction model trained on CIFAR-10.}
\label{fig:visual_analysis}
\vspace{-0.4cm}
\end{figure*}


\subsection{Data Prediction Model}
For experiments on data prediction models, we followed the configuration outlined by \citet{karras2022elucidating}. We apply the D-ODE solver to DDIM, rebuilt based on this configuration, and EDM~\citep{karras2022elucidating}, which employs Heun's second-order method. While \citet{karras2022elucidating} also re-implemented Euler-based samplers in their paper, we choose not to include them in our experiments, as EDM demonstrates superior FID scores.

\cref{fig:sample_subfigures} demonstrates that D-ODE solvers outperform ODE solvers, especially for smaller NFE. For instance, D-DDIM with 25 NFE can produce samples comparable to DDIM with 250 NFE in terms of FID, resulting in a speedup of around 10 times. With increasing NFE, FID scores of both ODE and D-ODE solvers asymptotically converge to each other. Given that the performance of student sampling is closely tied to that of teacher sampling, it is natural to observe similar FID scores for student and teacher sampling with larger NFE. Moreover, it is worth noting that around NFE 2, DDIM occasionally outperforms D-DDIM slightly. This observation suggests that the 2-step DDIM may not possess sufficient capacity to effectively distill knowledge from teacher sampling, particularly when DDIM is already generating noisy images (FID score exceeding 250).

\subsection{Comparison with Previous Distillation Methods}

The distillation process for D-ODE solvers typically requires only a few CPU minutes, adding negligible computational overhead to the entire sampling process. In contrast, previous distillation techniques for diffusion models~\citep{salimans2022progressive, meng2023distillation, song2023consistency} necessitate the optimization of the entire parameters of the denoising network. As a result, these methods demand a substantial amount of training time for each setting involving datasets, samplers, and networks.

\cref{tab:comparison_computational_cost} directly compares the computational times required by each distillation method to reach 3 FID on CIFAR-10 given the same pre-trained denoising network. The total time encompasses the distillation time following their configurations and the sampling time to generate 50k samples. For instance, D-EDM first optimizes $\lambda$ and then proceeds with the sampling process, while consistency distillation (CD)~\citep{song2023consistency} and progressive distillation (PD)~\citep{salimans2022progressive} need numerous training iterations before executing a few-step sampling.

\begin{table}
\centering
\begin{tabular}{lccc}
\toprule
    Method & D-EDM (Ours)   & CD~\cite{song2023consistency}  & PD~\cite{salimans2022progressive}   \\ \midrule
    Time   & $2.55$ & $187.25$ & $106.16$ \\
\bottomrule
\end{tabular}
\vspace{0.2cm}
\caption{\textbf{Comparison on computational time} to achieve 3 FID. The unit of time corresponds to the time required to generate 50k samples with 10-step DDIM.}
\label{tab:comparison_computational_cost}
\vspace{-0.4cm}
\end{table} 

The results clearly demonstrate that optimizing ODE solvers instead of the denoising network can significantly reduce computational time and resource requirements while achieving comparable sample quality. It is important to note that the results may vary depending on the training configuration of CD and PD, as the majority of their time is consumed during the distillation process. In this context, our method aligns well with the recent trend of democratizing diffusion models by minimizing or circumventing extensive training that relies on a large number of GPUs~\citep{hang2023efficient, wang2023patch, zheng2023fast, wu2023fast}. Supplementary Material provides a detailed explanation for distillation methods and additional comparisons with other sampling methods.

\section{Analysis}~\label{sec:analysis}
\vspace{-0.3cm}

This section encompasses visualizations of the sampling process and qualitative results. We initiate the exploration with a visual analysis following the methodology of \citet{liu2021pseudo}, aiming to scrutinize both global and local characteristics of the sampling process. Subsequently, we delve into a comparison of the generated images produced by ODE solvers and D-ODE solvers.

\begin{figure*}[h]
	\centering
	\begin{subfigure}{\linewidth}
        \centering
		\includegraphics[width=0.9\linewidth]{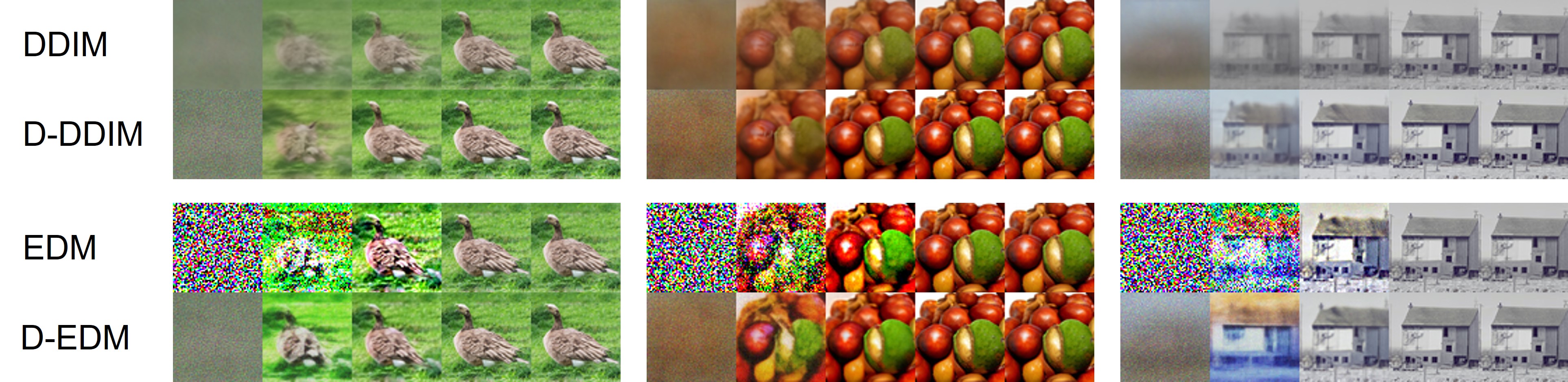}
		\caption{ImageNet $(64\times64)$}
		\label{fig:qualitative_subfigA}
	\end{subfigure}\\
        \vspace{1em}
	\begin{subfigure}{\linewidth}
            \centering
		\includegraphics[width=0.9\linewidth]{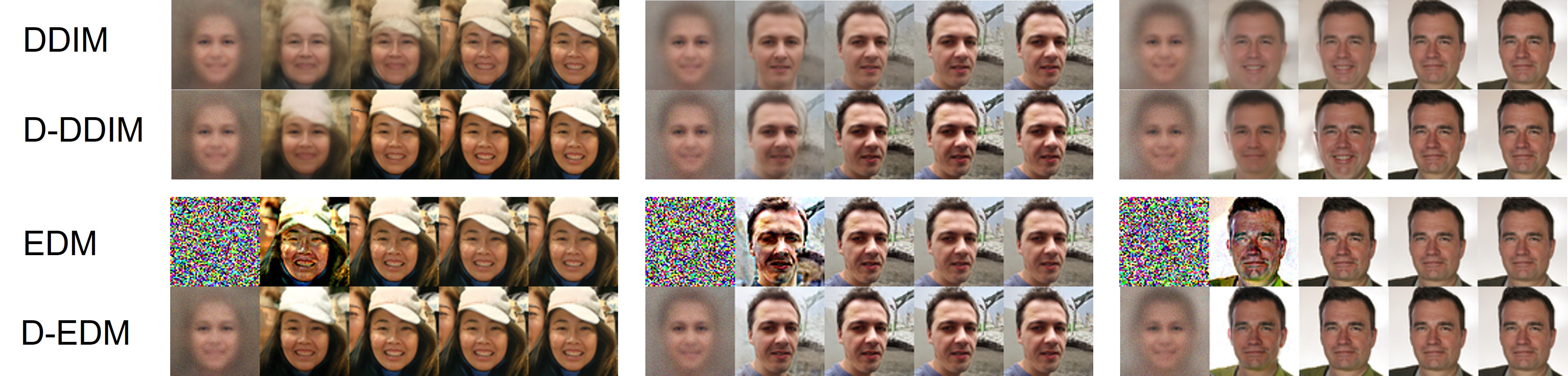}
		\caption{FFHQ $(64\times64)$}
		\label{fig:qualitative_subfigB}
	\end{subfigure}
	\caption{\textbf{Qualitative results.} Comparison of generated samples between ODE and D-ODE solvers. Data prediction models are used with increasing NFE (DDIM and D-DDIM: \{2, 5, 10, 25, 50\}, EDM and D-EDM: \{3, 5, 9, 25, 49\}). D-ODE solvers generate more realistic images compared to ODE solvers, especially for small NFE. }
	\label{fig:qualitative_subfigures}
 \vspace{-0.4cm}
\end{figure*}

\subsection{Visualization of Sampling Trajectory}

To facilitate the interpretation of high-dimensional data, we employ two distinct measures: the change in the norm as a global feature and the change in specific pixel values as a local feature, following the analysis scheme provided by \citet{liu2021pseudo}. For reference, the norm of DDIM with 1000 steps is included as it adheres to the target data manifold.

In the top row of \cref{fig:visual_analysis}, the norm of D-ODE solvers closely follows the trajectory traced by the norm of ODE solvers. This observation suggests that D-ODE solvers remain within the high-density regions of the data, exerting minimal influence on the ODE trajectory. This aligns with our design objective for D-ODE solvers, ensuring that the new denoising output matches the mean of the denoising output of ODE solvers, as discussed in \cref{subsec:formulation_d_ode_solver}.

In the bottom row of \cref{fig:visual_analysis}, two pixels are randomly selected from the image, and the change in their values is depicted, referencing the 1000-step DDIM as the target. Clearly, the pixel values of D-ODE solvers exhibit closer proximity to the target trajectory than those of ODE solvers. The results demonstrate that smaller-step D-ODE solvers can generate high-quality local features in samples, comparable to larger-step DDIM. It also emphasizes the importance of the data-specific parameter $\lambda$ to further reduce the fitting error of the score function. In conclusion, D-ODE solvers can achieve high-quality image generation by guiding their pixels toward the desired targets while remaining faithful to the original data manifold.

\subsection{Qualitative Analysis}

In \cref{fig:qualitative_subfigures}, we present a comparison of the generated images produced by ODE and D-ODE solvers using data prediction models trained on the ImageNet and FFHQ datasets. Generally, our method exhibits an improvement in image quality over ODE solvers, particularly for smaller NFE. DDIM tends to generate blurry images with indistinct boundaries, while D-DDIM produces clearer images with more prominent color contrast. EDM, especially with NFE smaller than 5, generates images characterized by high noise levels and artifacts, leading to FID scores exceeding 250. In contrast, D-EDM manages to generate relatively clear objects even at 5 NFE. Additional analysis figures and qualitative results can be found in the Supplementary Material.

\section{Conclusion}
\vspace{0.3cm}

In this study, we present D-ODE solvers, an innovative distillation method for diffusion models leveraging the principles of existing ODE solvers. Formulated by introducing a single parameter to ODE solvers, D-ODE solvers efficiently distill knowledge from teacher sampling with larger steps into student sampling with smaller steps, requiring minimal additional training. Our experiments showcase the efficacy of D-ODE solvers in enhancing the FID scores of state-of-the-art ODE solvers, especially in scenarios involving smaller NFE. Visual analyses provide insights into both global and local features of our method, revealing substantial improvements in image quality.

While the magnitude of improvement tends to be marginal or limited for large NFE values, eventually converging to the FID score of the teacher sampling process, D-ODE solvers remain an attractive option for augmenting sample quality with negligible additional computational cost. Their applicability extends across various samplers, datasets, and networks. However, for the generation of high-resolution images, the single-parameter nature of D-ODE solvers may prove insufficient. Exploring the incorporation of local-specific parameters, achieved through image grid divisions or latent space manipulations~\citep{rombach2022high}, presents an intriguing avenue for future research. 

\clearpage
\setcounter{page}{1}
\maketitlesupplementary

\section{Trilemma of Generative Models}~\label{sup_sec:trilemma}

Generative models face a trilemma characterized by three essential components, as outlined by \cite{xiao2021tackling}: 

\begin{enumerate}
    \item \textbf{High-quality samples}: Generative models should demonstrate the capacity to produce high-quality samples.
    \item \textbf{Mode coverage and sample diversity}: They ought to exhibit mode coverage, ensuring that generated samples are diverse and encompass various modes within the data distribution.
    \item \textbf{Fast sampling}: Efficient generative models should possess the ability to generate samples rapidly.
\end{enumerate}

For instance, generative adversarial networks (GANs)~\citep{goodfellow2014generative, brock2018large} excel in generating high-quality samples with just a single evaluation of the network. Nevertheless, GANs often struggle with generating diverse samples, resulting in poor mode coverage~\citep{salimans2016improved, zhao2018bias}. Conversely, Variational Autoencoders (VAEs)~\citep{kingma2013auto} and Normalizing Flows~\citep{dinh2016density} are designed to adequately ensure mode coverage but may suffer from low sample quality. Recently, diffusion models have emerged as a novel class of generative models that can generate high-quality samples comparable to GANs~\citep{dhariwal2021diffusion, saharia2022photorealistic}, while also providing a rich variety of samples. However, conventional diffusion models often require hundreds to thousands of network evaluations for sampling, rendering them computationally expensive in practice. The primary bottleneck in the sampling process of diffusion models is closely tied to the number of denoising network evaluations. Consequently, numerous research works have explored techniques to expedite the sampling process by either skipping or optimizing sampling steps while maintaining the quality of generated samples. These techniques can be broadly classified into two categories: learning-based and learning-free sampling methods~\citep{yang2022diffusion2} as introduced in the introduction of the main paper.

\section{Noise and Data Prediction Networks}~\label{sup_sec:noise_and_data_prediction_models}

The output of the denoising network should be parameterized to estimate the score function referring to the reverse-time ODE. The score function represents the gradient of the logarithm of the data distribution, indicating the direction of data with higher likelihood and less noise. One straightforward approach for the parameterization is to directly estimate the original data $\boldsymbol{x}$, in which case the score function is estimated by calculating the gradient toward the original data given the current noise level:
\begin{equation}
    \nabla_{\boldsymbol{x}} \log q_t(\boldsymbol{x}_t) = \frac{\boldsymbol{x}_{\boldsymbol{\theta}}(\boldsymbol{x}_t, t) - \boldsymbol{x}_t}{\sigma_t^2}.
\end{equation}
Another approach indirectly designs the denoising network to predict noise $\boldsymbol{\epsilon}$, which represents the residual signal infused in the original sample. In this case, the score function can be calculated as:
\begin{equation}
    \nabla_{\boldsymbol{x}} \log q_t(\boldsymbol{x}_t) = - \frac{\boldsymbol{\epsilon}_{\boldsymbol{\theta}}(\boldsymbol{x}_t, t)}{\sigma_t}.
\end{equation}
While the noise prediction network $\boldsymbol{\epsilon}_{\boldsymbol{\theta}}$ and the data prediction network $\boldsymbol{x}_{\boldsymbol{\theta}}$ are theoretically equivalent~\citep{kingma2021variational, luo2022understanding, karras2022elucidating}, they reveal different characteristics during the sampling process. 

\paragraph{Noise prediction network} Noise prediction networks may initially introduce significant discrepancies between the ground truth noise and the predicted noise~\citep{benny2022dynamic}. Since sampling commences with highly noisy samples, the denoising network lacks sufficient information to accurately predict noise~\cite {ho2020denoising}. Additionally, the magnitude of correction required at each timestep is relatively small, necessitating multiple timesteps to rectify such deviations~\citep{luo2022understanding}. 

\paragraph{Data prediction network} Data prediction networks are known to offer better accuracy in the initial stages of sampling, while the noise prediction networks become preferable in later stages. Predicting data assists the denoising network in understanding the global structure of the target sample~\citep{luo2022understanding}. Empirical evidence shows that the predicted data is close to the ground truth at the beginning of the sampling procedure~\citep{ramesh2022hierarchical, guan20223d}. However, in the later stages when substantial structures have already been formed and only minor noise artifacts need to be removed, finer details become challenging to recover~\citep{benny2022dynamic}. Essentially, the information provided by early data prediction becomes less effective in the later stages of sampling. 

\paragraph{Our experiments} The difference between data and noise prediction networks is also evident in the figure of the main paper, illustrating correlation between denoising outputs. Predictions of $\boldsymbol{\epsilon}$ in the initial sampling stages exhibit higher correlation with each other than those in later stages, whereas predictions of $\boldsymbol{x}$ become more correlated in the later stages compared to the earlier stages. In the case of noise estimation, a small amount of noise remains in a sample for the last few timesteps, resulting in detailed and minor changes to the sample with high variance. In conclusion, different details are modified at each timestep during the later sampling process. 

On the other hand, it is challenging for a $\boldsymbol{x}$ estimator to predict the original sample from the initial noisy sample. However, its predictions become more consistent in the later stages of sampling as the sample becomes less noisy. This observation aligns with the analysis presented in \citet{benny2022dynamic}, which indicates that the variance of the $\boldsymbol{x}$ estimator gradually decreases with more sampling steps, while the variance of the $\boldsymbol{\epsilon}$ estimator abruptly increases in the last phase of sampling.

\section{Knowledge distillation in Diffusion Models}~\label{sup_sec:distillation_in_diffusion_models}

Knowledge distillation~\citep{hinton2015distilling} was initially introduced to transfer knowledge from a larger model (teacher) to a smaller one (student), with the student model being trained to imitate the output of the teacher model. This concept can be applied to diffusion-based sampling processes to merge several timesteps (teacher) into a single timestep (student) to accelerate generation speed. 

\citet{luhman2021knowledge} directly apply knowledge distillation to diffusion models by minimizing the difference between the outputs of a one-step student sampler and the outputs of a multi-step DDIM sampler. Thus, the student model is trained to imitate the output of the teacher model, being initialized with a pre-trained denoising network to inherit knowledge from the teacher. 

Subsequently, progressive distillation~\citep{salimans2022progressive} proposes an iterative approach to train a student network to merge two sampling timesteps of the teacher network until it achieves one-step sampling to imitate the entire sampling process. This allows the student network to gradually learn the teacher's sampling process, as learning to predict the output of two-step sampling is easier than learning to predict the output of multi-step sampling. Given a pre-trained denoising network $\boldsymbol{\theta}$ as the teacher, \citet{salimans2022progressive} first train a student network $\boldsymbol{\theta}'$ to predict the output of two sampling timesteps of the teacher network. The student $\boldsymbol{\theta}'$ then becomes the new teacher and a new student with parameter $\boldsymbol{\theta}''$ is trained to combine two sampling timesteps of the new teacher network $\boldsymbol{\theta}'$ until the total timestep reaches one step. The student model is parameterized and initialized with the same deep neural network as the teacher model, and progressive distillation is examined with the DDIM sampler.  

\citet{meng2023distillation} extend progressive distillation to scenarios involving classifier-free diffusion guidance, achieving single-step or few-step generation for text-to-image generation, class-conditioned generation, image-to-image translation, and image inpainting. They leverage a two-stage approach to train a student model to match the combined output of the conditional and unconditional models first, and then apply progressive distillation by setting the student model as the new teacher. Most of the configuration remains the same as \citet{salimans2022progressive}, mainly utilizing DDIM sampler.

Recently, \citet{song2023consistency} proposes a new class of generative models called consistency models which exploit the consistency property on the trajectory of a probabilistic flow ODE. They are trained to predict the original sample from any point on the same ODE trajectory. During training, a target network and an online network are utilized so that the online network is optimized to generate the same output as the target network, while the target network is updated with an exponential moving average. Consistency models can generate samples in a single step or a few steps by design and are also capable of image inpainting, colorization, and super-resolution in a zero-shot fashion. They can be trained either independently or via distillation, which is named as consistency training and consistency distillation, respectively. In this paper, we are interested in consistency distillation in comparison with our distillation method.

However, these distillation methods typically require extensive training to adapt to different pre-trained models, datasets, and ODE solvers, which limits their practical applicability. In this paper, we propose to optimize newly parameterized ODE solvers (D-ODE solvers) exclusively. This approach effectively distills the sampling process with larger steps into a new process with smaller steps while keeping the pre-trained denoising network fixed. Because our method does not require parameter updates for the denoising network, the distillation process can be completed in just a few CPU minutes.

\section{Implementation Details of D-ODE Solvers}~\label{sup_sec:d_ode_solver}

In this section, we explain the ODE solvers of our interest in detail and their application in the framework of D-ODE solvers. We categorize ODE solvers into two distinct types based on the nature of the diffusion timestep: discrete and continuous. Discrete-time ODE solvers include DDIM, PNDM, DPM-Solver, and DEIS, where we built our code upon \citet{lu2022dpm}, while continuous-time ODE solvers contain re-implementations of DDIM and EDM based on the work done by \citet{karras2022elucidating}.

\subsection{D-ODE Solvers in Noise Prediction Networks}~\label{sup_subsec:d_ode_noise}

DDIM~\citep{song2020denoising} is formulated as a non-Markovian diffusion process of DDPM~\citep{ho2020denoising}, defining a deterministic generation procedure using implicit models. Given the estimated sample $\hat{\boldsymbol{x}}_t$ at timestep $t$, DDIM sampling process is expressed as follows:
\begin{equation}
    \hat{\boldsymbol{x}}_{t-1} = \alpha_{t-1}  \left(\frac{\hat{\boldsymbol{x}}_t-\sigma_t \boldsymbol{d}_t}{\alpha_t} \right) +  \sigma_{t-1} \boldsymbol{d}_t,
\end{equation}
where $\boldsymbol{d}_t = \boldsymbol{D}_{\boldsymbol{\theta}}(\hat{\boldsymbol{x}}_t, t)$ holds with the denoising network $\boldsymbol{D}_{\boldsymbol{\theta}}$.
Here, ($\alpha_t$, $\sigma_t$) represents a predefined noise schedule and the denoising network is parameterized as a noise prediction network $\boldsymbol{\epsilon}_{\boldsymbol{\theta}}$. The new denoising output $\boldsymbol{O}_t$, formulated by D-ODE solver, is defined as
\begin{equation}
   \boldsymbol{O}_t = \boldsymbol{d}_t + \lambda_t (\boldsymbol{d}_t - \boldsymbol{d}_{t+1}), 
\end{equation}
 following the notation in the main paper. We then simply substitute the denoising output $\boldsymbol{d}_t$ in the sampling process with the new one $\boldsymbol{O}_t$ :
\begin{equation}
    \hat{\boldsymbol{x}}_{t-1} = \alpha_{t-1} \left(\frac{\hat{\boldsymbol{x}}_t-\sigma_t \boldsymbol{O}_t}{\alpha_t} \right)+  \sigma_{t-1} \boldsymbol{O}_t.
    \label{supeq:d-ode_solver_update}
\end{equation}

Above equation defines D-DDIM sampling process with $\lambda_t$ to be optimized through knowledge distillation. In cases where the previous denoising output is unavailable (\eg, at timestep $T$), we use the given noisy sample to define new denoising output $\boldsymbol{O}_t$, resulting in $\boldsymbol{O}_t = \boldsymbol{d}_T + \lambda_T (\boldsymbol{d}_T - \boldsymbol{x}_T)$ at initial timestep $T$. The assumption that both $\boldsymbol{x}_T$ and $\boldsymbol{d}_T$ follow a normal distribution $\mathcal{N}(\boldsymbol{0}, \sigma_t^2 \boldsymbol{I})$ in theory ensures that the mean of $\boldsymbol{O}_t$ remains consistent with the original denoising output. It is expected that $(\boldsymbol{d}_T - \boldsymbol{x}_T)$ contains information regarding the direction toward the true $\boldsymbol{x}_{T-1}$ to some extent, which actually improves the FID score in practice. Thus, we also apply this sampling recipe to other D-ODE solvers based on noise prediction networks.

PNDM~\citep{liu2021pseudo} is based on pseudo-numerical methods on the data manifold, built upon the observation that classical numerical methods can deviate from the high-density area of data. PNDM encapsulates DDIM as a simple case and surpasses DDIM with its high-order methods. However, PNDM requires 12 NFE for the first 3 steps, making it challenging to compare with other methods using a fixed NFE. Therefore, we opt for iPNDM~\citep{zhang2022fast}, which eliminates the need for initial warm-up steps and outperforms PNDM while maintaining the pseudo-numerical sampling process. iPNDM employs a linear combination of multiple denoising outputs to represent the current denoising output while adhering to the sampling update path of DDIM, as shown below:
\begin{gather}
    \hat{\boldsymbol{d}}_t^{(3)} = \frac{1}{24}(55\boldsymbol{d}_t - 59\boldsymbol{d}_{t+1}+37\boldsymbol{d}_{t+2}-9\boldsymbol{d}_{t+3} ),   \\
    \hat{\boldsymbol{x}}_{t-1} = \alpha_{t-1}  \left(\frac{\hat{\boldsymbol{x}}_t-\sigma_t \hat{\boldsymbol{d}}_t^{(3)}}{\alpha_t} \right) +  \sigma_{t-1} \hat{\boldsymbol{d}}_t^{(3)},\label{supeq:ipndm_update}  
\end{gather}
where $\hat{\boldsymbol{d}}_t$ is approximated with three previous denoising outputs (\ie, $\boldsymbol{d}_{t+1}$, $\boldsymbol{d}_{t+2}$, and $\boldsymbol{d}_{t+3}$) and then applied to the DDIM sampling process. Therefore, the first three denoising outputs should be defined independently as follows:
\begin{align}
    \hat{\boldsymbol{d}}_t^{(0)} &= \hat{\boldsymbol{d}}_t, \\
    \hat{\boldsymbol{d}}_t^{(1)} &= \frac{3}{2}\hat{\boldsymbol{d}}_t -\frac{1}{2}\hat{\boldsymbol{d}}_{t+1}, \\
    \hat{\boldsymbol{d}}_t^{(2)} &= \frac{1}{12}(23\hat{\boldsymbol{d}}_{t} -16\hat{\boldsymbol{d}}_{t+1} +5\hat{\boldsymbol{d}}_{t+2}).
\end{align}
Leveraging these newly defined denoising outputs $\hat{\boldsymbol{d}}_t^{(p)}$ ($p=3$ after three timesteps) by iPNDM, we construct the sampling process of D-iPNDM, where the new denoising output $\boldsymbol{O}_t$ can be defined as 
\begin{equation}
  \boldsymbol{O}_t = \hat{\boldsymbol{d}}_t^{(p)} + \lambda_t (\hat{\boldsymbol{d}}_t^{(p)} - \hat{\boldsymbol{d}}_{t+1}^{(p)}).  
\end{equation}
Then, $\hat{\boldsymbol{d}}_t^{(p)}$ in \cref{supeq:ipndm_update} is replaced by $\boldsymbol{O}_t$, which leads to the same update rule as \cref{supeq:d-ode_solver_update} with differrent formulation of $\boldsymbol{O}_t$.

DPM-Solver~\citep{lu2022dpm} utilizes the semi-linear structure of probabilistic flow ODEs by solving the exact formulation of the linear part of ODEs and approximating the weighted integral of the neural network with exponential integrators~\citep{hochbruck2010exponential}. DPM-Solver offers first-order, second-order, and third-order methods, with the first-order variant corresponding to DDIM. For single step approach, DPM-Solver strategically divides the total sampling steps using these different-order methods. For instance, DPM-Solver2 (second-order DPM-Solver) is employed 5 times to generate a sample comprising 10 denoising steps, with the denoising network being evaluated twice within DPM-Solver2. To achieve 15 denoising steps, DPM-Solver2 is applied 7 times, and DPM-Solver1 (or DDIM) is applied during the final denoising step.

In this section, we delve into the formulation of D-DPM-Solver2, and the application to DPM-Solver3 and DPM-Solver++~\cite{lu2022dpm++} follows a similar approach. First, we denote $\tau_t = \log (\alpha_t/\sigma_t)$ as the logarithm of the signal-to-noise ratio (SNR), and $\tau_t $ is a strictly decreasing function as $t$ increases. Consequently, we can establish an inverse function mapping from $\tau$ to $t$, denoted as $t_\tau (\cdot) : \mathbb{R} \rightarrow \mathbb{R}$. Now, we can outline DPM-Solver2 with the following steps:
\begin{align}
    t-\frac{1}{2} &= t_\tau(\frac{\tau_{t-1}+\tau_{t}}{2}),  \\
    \hat{\boldsymbol{x}}_{t-\frac{1}{2}} &= \frac{\alpha_{t-\frac{1}{2}}}{\alpha_{t}} \hat{\boldsymbol{x}}_t - \sigma_{t-\frac{1}{2}} (e^{\frac{h_t}{2}}-1) \boldsymbol{d}_t, \label{supeq:first_step_dpmsolver2} \\
    \hat{\boldsymbol{x}}_{t-1} &= \frac{\alpha_{t-1}}{\alpha_{t}}\hat{\boldsymbol{x}}_t - \sigma_{t-1}(e^{h_t}-1)\boldsymbol{d}_{t-\frac{1}{2}}. \label{supeq:second_step_dpmsolver2}
\end{align}
In these equations, $h_t = \tau_{t-1}-\tau_t$, and $\hat{\boldsymbol{x}}_{t-\frac{1}{2}}$ represents the intermediate output between timestep $t-1$ and $t$. Since DPM-Solver2 utilizes a two-stage denoising step, we must define two denoising outputs $\boldsymbol{O}_t$ and $\boldsymbol{O}_{t-\frac{1}{2}}$ to formulate D-DPM-Solver2 with $\lambda_t$ and $\lambda_{t-\frac{1}{2}}$ optimized through knowledge distillation:
\begin{align}
 \boldsymbol{O}_t &= \boldsymbol{d}_t + \lambda_t (\boldsymbol{d}_t - \boldsymbol{d}_{t+\frac{1}{2}}), \\
 \boldsymbol{O}_{t-\frac{1}{2}} &= \boldsymbol{d}_{t-\frac{1}{2}} + \lambda_{t-\frac{1}{2}} (\boldsymbol{d}_{t-\frac{1}{2}} - \boldsymbol{d}_t).
\end{align}
These new denoising outputs are then applied in \cref{supeq:first_step_dpmsolver2} and \cref{supeq:second_step_dpmsolver2} to define the sampling process of D-DPM-Solver2:
\begin{align}
    \hat{\boldsymbol{x}}_{t-\frac{1}{2}} &= \frac{\alpha_{t-\frac{1}{2}}}{\alpha_{t}} \hat{\boldsymbol{x}}_t - \sigma_{t-\frac{1}{2}} (e^{\frac{h_t}{2}}-1) \boldsymbol{O}_t,\\
    \hat{\boldsymbol{x}}_{t-1} &= \frac{\alpha_{t-1}}{\alpha_{t}}\hat{\boldsymbol{x}}_t - \sigma_{t-1}(e^{h_t}-1)\boldsymbol{O}_{t-\frac{1}{2}}. 
\end{align}

Similar to DPM-Solver, DEIS~\citep{zhang2022fast} employs an exponential integrator to exploit the semi-linear structure of the reverse-time diffusion process. In particular, they propose the use of high-order polynomials to approximate the non-linear term in ODEs as shown below:
\begin{align}
    P_r(t) &= \sum_{j=0}^r C_{tj} \boldsymbol{d}_{t+j}, \\
    \hat{\boldsymbol{x}}_{t-1} &= \frac{\alpha_{t-1}}{\alpha_{t}}\hat{\boldsymbol{x}}_{t} + P_r(t),\label{supeq:deis_formulation} 
\end{align}

where $\{C_{tj}\}_{j=0}^r$ is numerically determined through weighted integration to approximate the true ODE trajectory. DEIS offers several variants based on the numerical method used to estimate $C_{tj}$, and for our experiments, we choose $t$AB-DEIS as it exhibits the most promising results among the variants. Additionally, \citet{zhang2022fast} explores DEIS for different values of $r \in \{1, 2, 3\}$ where larger values of $r$ generally lead to improved approximations of the target score function. It is worth noting that DDIM can be seen as a special case of $t$AB-DEIS with $r=0$.

Referring to \cref{supeq:deis_formulation}, we define a new denoising output $\boldsymbol{O}_t$ and the sampling process of D-DEIS as follows:
\begin{align}
   \boldsymbol{O}_t &=  P_r(t) + \lambda_t (P_r(t) - P_r(t+1)), \\
    \hat{\boldsymbol{x}}_{t-1} &= \frac{\alpha_{t-1}}{\alpha_{t}}\hat{\boldsymbol{x}}_{t} + \boldsymbol{O}_t. 
\end{align}

\subsection{D-ODE Solvers in Data Prediction Networks}~\label{sup_subsec:d_ode_data}

In our study, we newly implement DDIM~\citep{song2020denoising} in a continuous setting using the parameterization of the data prediction network. We follow the configurations outlined by \citet{karras2022elucidating}. The sampling process for this modified DDIM is defined as follows:
\begin{align}
   s_t &= \frac{\boldsymbol{d}_t - \hat{\boldsymbol{x}}_t}{\sigma_t}, \\
\hat{\boldsymbol{x}}_{t-1} &= \hat{\boldsymbol{x}}_t + (\sigma_t - \sigma_{t-1}) s_t, \label{supeq:edm_ddim_update_step}
\end{align}
where $\boldsymbol{d}_t = \boldsymbol{D}_{\boldsymbol{\theta}}(\hat{\boldsymbol{x}}_t, t)$ holds, and $s_t$ approximates the score function, directing toward the high-density area of the data. The denoising network is parameterized as the data prediction network $\boldsymbol{x}_{\boldsymbol{\theta}}$, and the denoising step is carried out in \cref{supeq:edm_ddim_update_step} based on the difference in noise levels measured by $(\sigma_t - \sigma_{t-1})$. 

Similar to D-DDIM with the noise prediction network, the new denoising output $\boldsymbol{O}_t$ for D-DDIM is defined as 
\begin{equation}
\boldsymbol{O}_t = \boldsymbol{d}_t + \lambda_t(\boldsymbol{d}_t-\boldsymbol{d}_{t+1}).    
\end{equation}
Then, $\boldsymbol{O}_t$ is incorporated into the sampling process of DDIM instead of $\boldsymbol{d}_t$ as follows:
\begin{align}
   s_t &= \frac{\boldsymbol{O}_t - \hat{\boldsymbol{x}}_t}{\sigma_t}, \\
\hat{\boldsymbol{x}}_{t-1} &= \hat{\boldsymbol{x}}_t + (\sigma_t - \sigma_{t-1}) s_t.
\end{align}

\citet{karras2022elucidating} introduce EDM sampler based on Heun's second-order method, which achieves a state-of-the-art FID score on CIFAR-10 and ImageNet64. They utilize a novel ODE formulation, parameter selection, and modified neural architectures. The EDM sampling process is shown as follows:
\begin{align}
   s_t &= \frac{\boldsymbol{d}_t - \hat{\boldsymbol{x}}_t}{\sigma_t}, \label{supeq:edm_update1} \\ \hat{\boldsymbol{x}}_{t-1}' &= \hat{\boldsymbol{x}}_t + (\sigma_t - \sigma_{t-1}) s_t, \label{supeq:edm_update2}\\
    s_t' &= \frac{\boldsymbol{d}_{t-1}' - \hat{\boldsymbol{x}}_{t-1}'}{\sigma_{t-1}},\label{supeq:edm_update3} \\ 
    \hat{\boldsymbol{x}}_{t-1} &= \hat{\boldsymbol{x}}_t + (\sigma_t - \sigma_{t-1}) (\frac{1}{2}s_t + \frac{1}{2}s_t'),\label{supeq:edm_update4}
\end{align}
where $\boldsymbol{d}_{t-1}' = \boldsymbol{D}_{\boldsymbol{\theta}}(\hat{\boldsymbol{x}}_{t-1}', t-1)$ holds. The first stage of EDM with \cref{supeq:edm_update1} and \cref{supeq:edm_update2} is equivalent to DDIM, and then the score function is more accurately estimated in the second stage with \cref{supeq:edm_update3} and \cref{supeq:edm_update4} by linearly combining two estimations $s_t$ and $s_t'$. Notably, 18 steps of EDM sampling correspond to 35 NFE, as one step of EDM involves two network evaluations, and \cref{supeq:edm_update3} and \cref{supeq:edm_update4} are not computed at the last step. 

To construct the sampling process of D-EDM, we define two denoising outputs: 
\begin{align}
\boldsymbol{O}_t &= \boldsymbol{d}_t + \lambda_t(\boldsymbol{d}_t-\boldsymbol{d}_{t+1}'), \\
\boldsymbol{O}_{t-1}' &= \boldsymbol{d}_{t-1}' + \lambda_t(\boldsymbol{d}_{t-1}'-\boldsymbol{d}_{t}).
\end{align}
Consequently, the sampling steps for D-EDM are described as follows:
\begin{align}
   s_t &= \frac{\boldsymbol{O}_t - \hat{\boldsymbol{x}}_t}{\sigma_t}, \\ \hat{\boldsymbol{x}}_{t-1}' &= \hat{\boldsymbol{x}}_t + (\sigma_t - \sigma_{t-1}) s_t, \\
    s_t' &= \frac{\boldsymbol{O}_{t-1}' - \hat{\boldsymbol{x}}_{t-1}'}{\sigma_{t-1}}, \\ 
    \hat{\boldsymbol{x}}_{t-1} &= \hat{\boldsymbol{x}}_t + (\sigma_t - \sigma_{t-1}) (\frac{1}{2}s_t + \frac{1}{2}s_t').
\end{align}

\subsection{Various Interpretations of D-ODE Solvers}
New denoising output $\boldsymbol{O}_t$ in D-ODE solvers is formulated based on the observation that denoising outputs are highly correlated, and it is essential to retain the same mean as the original outputs. We rewrite the definition of our denoising output as follows:
\begin{equation}
\boldsymbol{O}_t = \boldsymbol{d}_t + \lambda_t(\boldsymbol{d}_t-\boldsymbol{d}_{t+1}). \label{supeq:d-ode_formulation}
\end{equation}

The above formulation can be interpreted to calculate interpolation (or extrapolation) between the current and previous denoising outputs to estimate the accurate score function. Therefore, D-ODE solvers can be seen as the process of dynamically interpolating (or extrapolating) the denoising outputs with $\lambda_t$ optimized through knowledge distillation. Similarly, \citet{zhang2023lookahead} proposed the use of extrapolation on the current and previous estimates of the original data $\hat{\boldsymbol{x}}_t$. They argued that extrapolating between two predictions includes useful information toward the target data by refining the true mean estimation. Although accurate extrapolation requires grid search for parameter tuning, they demonstrated improvements in the FID of various ODE solvers.

Another interpretation is based on the work of \citet{permenter2023interpreting}, who matched the denoising process to gradient descent applied to the Euclidean distance function under specific assumptions. They reinterpreted diffusion models using the definition of projection onto the true data distribution and proposed a new sampler by minimizing the error in predicting $\boldsymbol{\epsilon}$ between adjacent timesteps. Their sampler corresponds to D-DDIM with $\lambda_t=1$ selected via grid search, and it outperforms DDIM and PNDM.

The last interpretation is that D-ODE solvers accelerate the convergence of sample generation in a way similar to how momentum boosts optimization in SGD~\citep{sutskever2013importance}. Just as SGD with momentum utilizes the history of previous gradients to speed up parameter updates in a neural network, D-ODE solvers leverage previous denoising outputs to accelerate the convergence of sampling. An interesting future direction could explore whether advanced optimizers used in machine learning models~\citep{kingma2014adam, duchi2011adaptive, ruder2016overview} can be effectively applied to diffusion models.

\subsection{Various Formulations of D-ODE Solvers} ~\label{sup_subsec:various_formulation_d_ode}

To further validate the effectiveness of D-ODE solvers, we explore different formulations of D-ODE solvers based on DDIM. For example, we can estimate parameters for two adjacent denoising outputs separately instead of optimizing a single parameter $\lambda_t$, which we name D-DDIM-Sep. D-DDIM-Sep corresponds to Eq.\ (8) of the main paper with $T=t+1$. Eq.\ (8) of the main paper is represented as D-DDIM-All where all previous denoising outputs are utilized to estimate the new one. Additionally, we include D-DDIM which is shown as Eq.\ (10) of the main paper and D-DDIM-2 which is equal to Eq.\ (9) of the main paper with $T=t+2$.
All methods are explicitly presented below for comparison, with $\boldsymbol{d}_t = \boldsymbol{D}_{\boldsymbol{\theta}}(\hat{\boldsymbol{x}}_t, t)$:
\begin{align}
    &\text{DDIM}: \hspace{0.5em} \boldsymbol{d}_t, \\
    &\text{D-DDIM-Sep}: \hspace{0.5em} \boldsymbol{O}_t= \lambda_{t1}\boldsymbol{d}_t+ \lambda_{t2}\boldsymbol{d}_{t+1}, ~\label{supeq:d_ddim_sep} \\
    &\text{D-DDIM-All}: \hspace{0.5em} \boldsymbol{O}_t= \sum_{k=t}^T \lambda_{k}\boldsymbol{d}_k, ~\label{supeq:d_ddim_all}\\
    &\text{D-DDIM}: \hspace{0.5em} \boldsymbol{O}_t= \boldsymbol{d}_t + \lambda_{t1}(\boldsymbol{d}_t-\boldsymbol{d}_{t+1}) ,\\
    \begin{split}
    \text{D-DDIM-2}: \hspace{0.5em} \boldsymbol{O}_t= \boldsymbol{d}_t + \lambda_{t1}(\boldsymbol{d}_t-\boldsymbol{d}_{t+1})+ \\ \lambda_{t2}(\boldsymbol{d}_t-\boldsymbol{d}_{t+2}).
    \end{split}
\end{align}

\begin{table}[h]
     \centering
    \begin{tabular}{lccc}
    \Xhline{2\arrayrulewidth}
    NFE & 10   & 25   & 50   \\ \hline\hline
    DDIM      & 18.85 & 9.79 & 7.17 \\
    D-DDIM-Sep     & 79.21 & 26.40 & 11.50 \\
    D-DDIM-All     & 179.67 & 36.65 & 18.48 \\
    D-DDIM        & \textbf{8.67} & \textbf{8.18} & \textbf{6.55} \\
    D-DDIM-2  & 18.75 & 9.83 & 7.21 \\ \Xhline{2\arrayrulewidth}
    \end{tabular}
    \vspace{0.5em}
     \caption{\textbf{Comparison on various D-ODE solver formulations.} FID is measured on CIFAR-10 with the noise prediction model and the best FID is bolded.}
    \label{suptab:comparison_d_ode_formulations}
\end{table}

\begin{figure}[h]
\begin{center}
\includegraphics[width=0.8\linewidth]{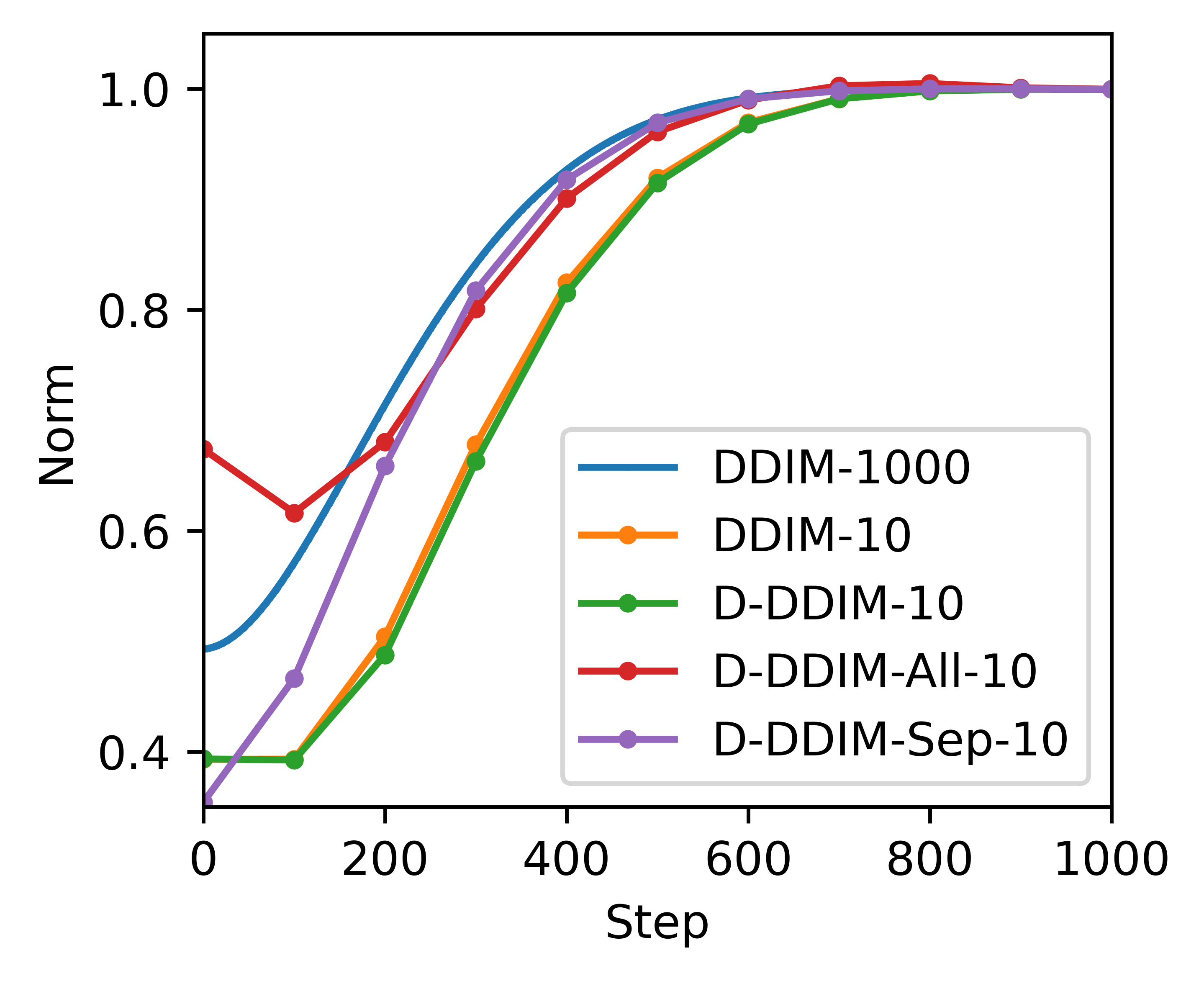} 
\end{center}
\caption{\textbf{Comparison of the change of norm with different formulations.} We adopt the same setting as Fig.\ 5 in the main paper.} \label{supfig:trajectory_comparison}
\end{figure}

We examined the five formulations mentioned above on CIFAR-10 with different NFE, while all other configurations for distillation and sampling are remained the same. As shown in \cref{suptab:comparison_d_ode_formulations}, D-DDIM outperforms all other formulations, and other formulations such as D-DDIM-Sep, D-DDIM-All, and D-DDIM-2 even worsen the FID score compared to DDIM. D-DDIM-Sep and D-DDIM-All results in especially higher FID scores which can be interpreted that the sampling process does not properly converge to generate realistic samples. As we pointed out in the main paper, independently estimated parameters may deviate from the target trajectory of ODE solvers. This is due to the fact that the set of $\lambda$ in \cref{supeq:d_ddim_sep} and \cref{supeq:d_ddim_all}, determined by distillation, can be volatile without any constraints and may not reflect the general sampling rules across different batches. D-DDIM-2 also does not improve the FID score of DDIM. One possible reason for this is that parameters optimized on one batch may not be applicable to others. Since the two parameters are optimized on only one batch, fine-grained estimation of denoising predictions like D-DDIM-2 may not be valid for all batches.

Moreover, we display the change of norm in \cref{supfig:trajectory_comparison} referring to Fig.\ 5 of the main paper. While D-DDIM-All-10 and D-DDIM-Sep-10 initially seem to follow the target trajectory (\ie, DDIM-1000), they highly deviate from either the target or the original ODE trajectory (\ie, D-DDIM-10) at last, which matches with the high FID scores in \cref{suptab:comparison_d_ode_formulations}. As mentioned in Sec.\ 3.2 this is due to the instability inherent in \cref{supeq:d_ddim_all}.

\section{Experiment Details}~\label{sup:experiment_details}

\paragraph{Model architectures} 
For the noise prediction models, we follow the architectures and configurations of \citet{ho2020denoising} and \citet{dhariwal2021diffusion}, utilizing their pre-trained models. Specifically, we adopt the model architecture and configuration in DDPM~\citep{ho2020denoising} for experiments on CIFAR-10 and CelebA $64\times64$. For ImageNet $128\times128$ and LSUN Bedroom $256\times256$, we use the corresponding network architectures from \citet{dhariwal2021diffusion}. In experiments with the data prediction models, we utilize the configurations and pre-trained models from \citet{karras2022elucidating} for CIFAR-10, FFHQ $64\times64$, and ImageNet $64\times64$.

\paragraph{Distillation configurations} 
As outlined in the algorithm of main paper, we first perform teacher sampling with $CT$ steps to set target samples, followed by student sampling with $T$ steps to match the student's outputs with the teacher's targets. For most D-ODE solvers, we use DDIM sampling as the teacher sampling method, as it generates one denoising output per denoising step, enabling one-to-one matching between targets and predictions. For iPNDM and DEIS, we use themselves as the teacher method for distillation, respectively (\eg, DEIS with $CT$ steps as the teacher and D-DEIS with $T$ steps as the student). While they use a linear combination of previous denoising outputs to estimate current denoising predictions, the sampling dynamics are the same as DDIM. Therefore, the teacher's targets and student's predictions can be easily matched. 

Moreover, student sampling is performed sequentially to optimize $\lambda$ in D-ODE solvers. In other words, $\lambda_t$ is first estimated via distillation and then the next sample at timestep $t+1$ is generated with optimized D-ODE solvers at timestep $t$ during student sampling. This approach helps stabilize the sampling process, as $\lambda_{t+1}$ is estimated based on previously generated samples from D-ODE solvers with $\lambda_t^*$. As a result, it can alleviate exposure bias~\citep{ranzato2016sequence, ning2023input} with precisely estimated $\lambda$.

\paragraph{Sampling details} 
For simplicity, we adopt uniformly divided timesteps for all ODE solvers. We generate 50K samples and report the mean FID score calculated after three runs with different seeds. All experiments are conducted using GPUs, including NVIDIA TITAN Xp, Nvidia V100, and Nvidia A100. We fix the scale $C=10$ and batch size $|B|=100$, except for LSUN Bedroom where $|B|=25$ due to memory constraints. Ablation studies on these two parameters are presented in \cref{sup:ablation_studies}. 

Several design choices need to be made for each ODE solver. PNDM requires 12 NFE for the first 3 steps, making it challenging to compare with other methods using a fixed NFE. Therefore, we adopt iPNDM~\citep{zhang2022fast}, which does not require initial warm-up steps and outperforms PNDM. DEIS offers various versions of ODE solvers, among which we select $t$AB-DEIS, exhibiting the best FID score in their experiments. DPM-Solver combines different-order solvers using adaptive step sizes. For simplicity, we opt for the single-step DPM-Solver, which sequentially uses DPM-Solver1, DPM-Solver2, and DPM-Solver3 to compose the total timesteps. While EDM allows stochastic sampling by its design, we employ deterministic sampling to obtain a definite target sample generated by teacher sampling.

\section{Ablation Studies}~\label{sup:ablation_studies}

\renewcommand{\arraystretch}{1.3}

\begin{table}[h]
    \centering
    \begin{subtable}[h]{0.45\textwidth}
        \centering 
        \begin{tabular}{lccc}
        \Xhline{2\arrayrulewidth}
        NFE & 10   & 25   & 50   \\ \hline \hline
        5                        & $9.68_{\pm 0.10}$ & $8.20_{\pm 0.06}$ & $6.52_{\pm 0.02}$ \\
        10                       & $8.83_{\pm 0.10}$ & $8.09_{\pm 0.03}$ & $6.55_{\pm 0.09}$ \\
        20                       & $8.52_{\pm 0.04}$ & $8.01_{\pm 0.03}$ & $\textbf{6.50}_{\pm 0.01}$ \\
        30                       & $\textbf{8.41}_{\pm 0.05}$ & $\textbf{7.87}_{\pm 0.05}$ & $\textbf{6.50}_{\pm 0.01}$ \\ \Xhline{2\arrayrulewidth}
        \end{tabular}
       \caption{Different Scale $S$}
       \label{subtab:ablation_scale}  
    \end{subtable}
    \hfill
    \begin{subtable}[h]{0.45\textwidth}
        \centering      
        \begin{tabular}{lccc}
        \Xhline{2\arrayrulewidth}
        NFE & 10   & 25   & 50   \\ \hline \hline
        5                        & $9.33_{\pm 0.66}$ & $7.75_{\pm 0.13}$ & $6.64_{\pm 0.09}$ \\
        10                       & $8.83_{\pm 0.58}$ & $7.79_{\pm 0.09}$ & $6.55_{\pm 0.07}$ \\
        50                       & $\textbf{8.03}_{\pm 0.08}$ & $7.69_{\pm 0.08}$ & $6.58_{\pm 0.05}$ \\
        100                       & $8.22_{\pm 0.10}$ & $\textbf{7.68}_{\pm 0.03}$ & $\textbf{6.50}_{\pm 0.09}$ \\ \Xhline{2\arrayrulewidth}
        \end{tabular}
        \caption{Different Batch Size $|B|$}
        \label{subtab:ablation_batch_size}  
     \end{subtable}
     \vspace{0.5em}
     \caption{\textbf{Ablation studies on scale $C \in \{5, 10, 20, 30\}$ and batch size $|B| \in \{5,10,50,100\}$.} CIFAR-10 with noise prediction models are employed for evaluation. We report mean and standard deviation after 3 runs (mean $\pm$ std) and the best FID is bolded.}
     \label{tab:ablation}
\end{table}

We conduct ablation studies on two key parameters for the distillation of D-ODE solvers: the scale $S$ and the batch size $|B|$. The scale $S$ determines the number of steps for the teacher sampling, with the teacher sampling going through $S$ times more denoising steps compared to the student sampling. A larger scale $S$ results in a better target generated by the teacher sampling and can be viewed as increasing the guidance strength of the teacher during distillation. It is also crucial to choose an appropriate batch size $|B|$ since the optimal $\lambda$ is estimated on a single batch $B$ and then reused for other batches. Thus, the batch size should be large enough to encompass different modes of samples within the dataset, while excessively large batch size may not fit into GPU memory.

We test various scales in \cref{subtab:ablation_scale} using the noise prediction models trained on CIFAR-10. As the scale increases, the FID score consistently improves across different NFE values. With a larger scale $S$, the student sampling is strongly guided by the accurate target of teacher sampling, resulting in a lower FID. However, the effect of the guidance scale weakens with increasing NFE. This is reasonable since the performance of student sampling depends heavily on that of teacher sampling, and the teacher's FID score eventually converges to a certain value. As the maximum number of timesteps is 1000 for discrete timesteps, scales 20 and 30 at 50 NFE generate samples guided by the same teacher sampling.

\begin{figure*}[h]
	\centering
	\begin{subfigure}[t]{0.3\linewidth}
	        \includegraphics[width=\linewidth]{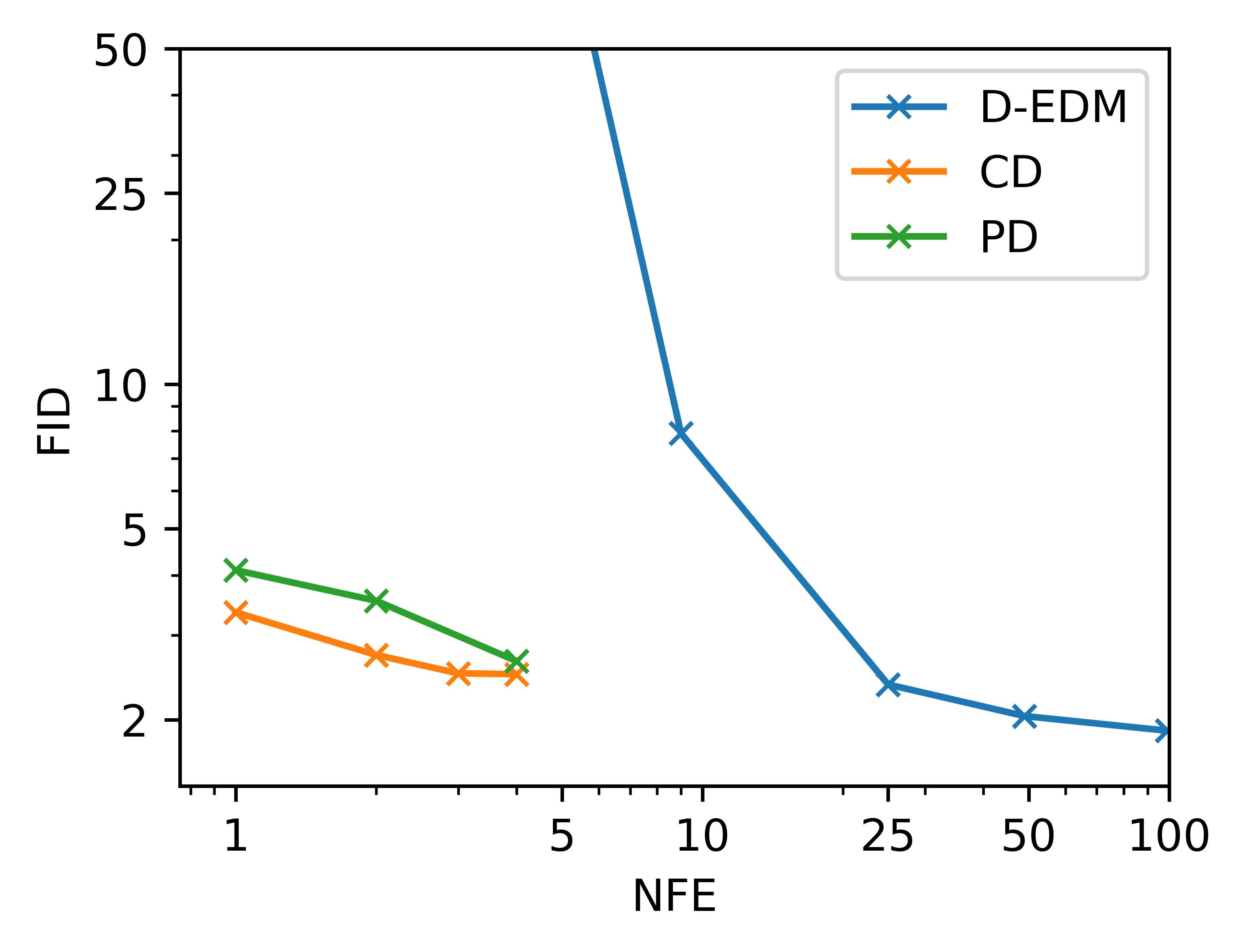}
	        \caption{Distillation methods}
            \label{supfigA:comparison_distillation}
         \end{subfigure}
	\begin{subfigure}[t]{0.3\linewidth}
		\includegraphics[width=\linewidth]{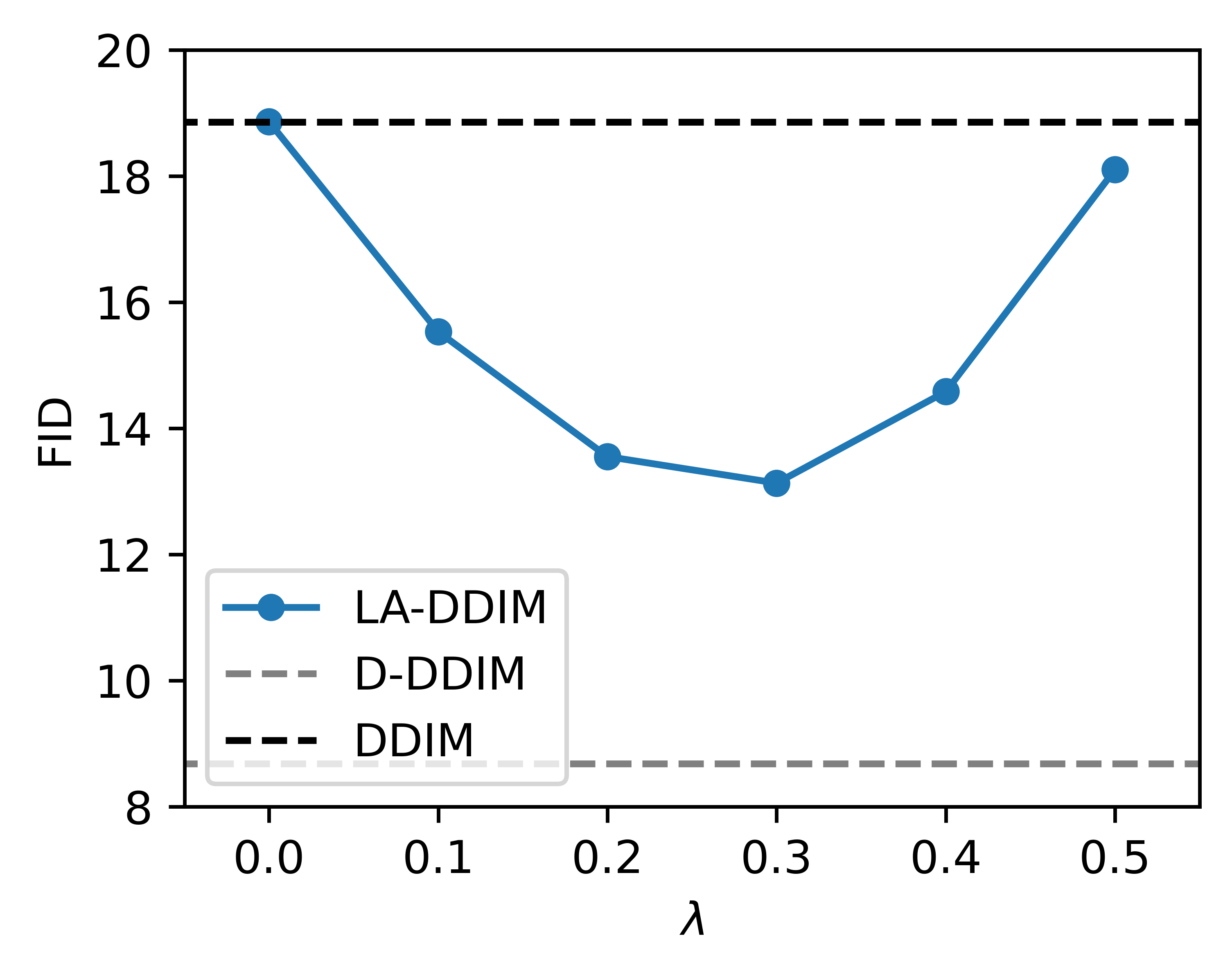}
		\caption{LA-DDIM}
         \label{supfigB:comparison_LA}
	\end{subfigure}
	\begin{subfigure}[t]{0.3\linewidth}
		\includegraphics[width=\linewidth]{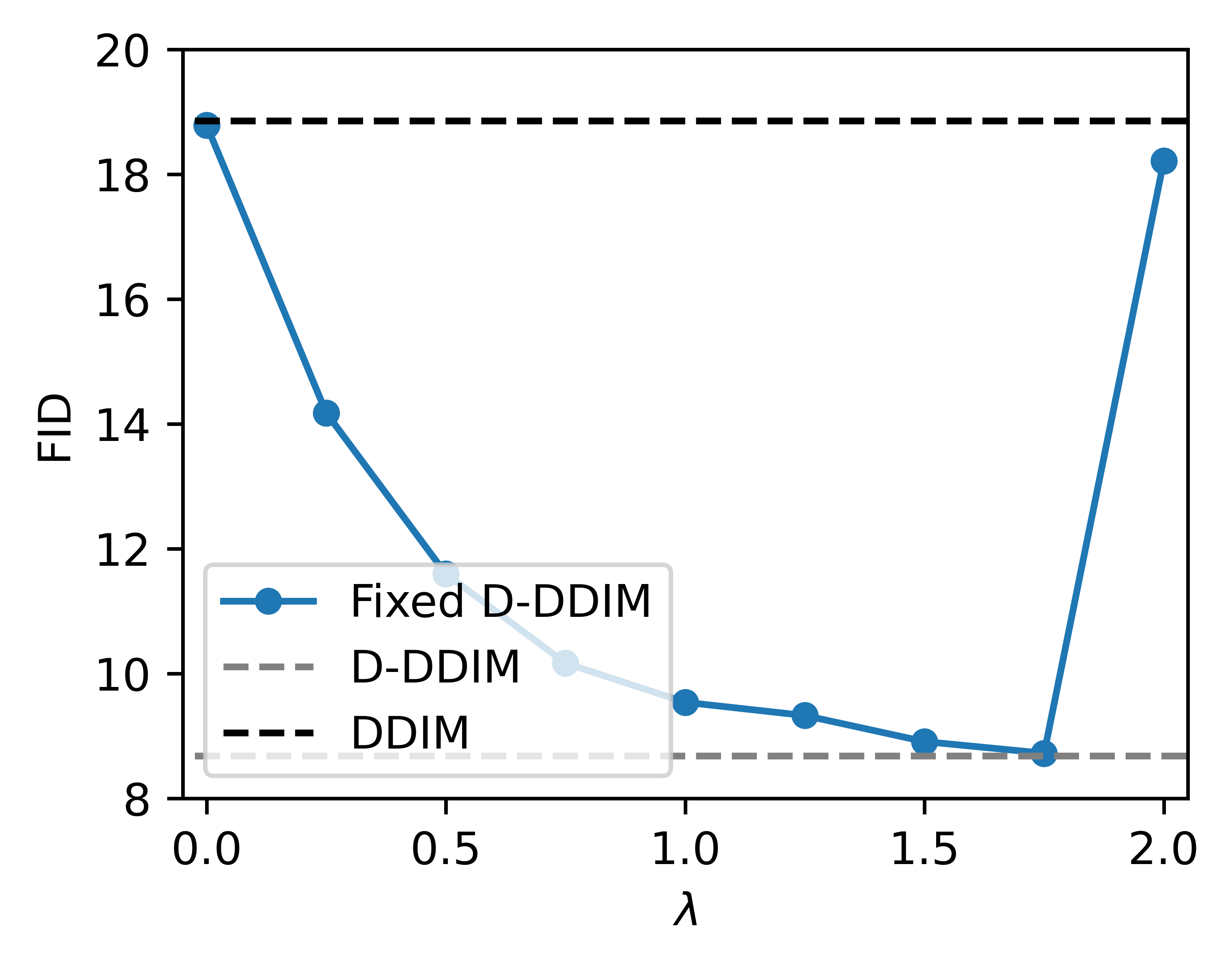}
	\caption{Fixed-D-DDIM}
        \label{supfigC:comparison_Fixed_d_ddim}
	\end{subfigure}
	\caption{\textbf{Comparison figures.} (a) FID scores over NFE for distillation methods (CD, PD, and D-EDM). (b) FID scores over $\lambda$ with LA-DDIM. (c) FID scores over $\lambda$ with Fixed-D-DDIM }
\end{figure*}


In \cref{subtab:ablation_batch_size}, D-ODE solvers with various batch sizes also exhibit clear tendency. As the batch size increases, both the FID score and variance tend to decrease. With relatively large NFE values, FID scores and variance converge to a certain point. As the effect of distillation diminishes with higher NFE, even a small batch size results in low variance. We choose a batch size of 100 for most datasets, which is sufficient to capture the inherent variety of the dataset and reduce variance compared to a smaller batch size.

\section{More Comparisons}~\label{sup:comparison}

\begin{table}[h]
     \centering
    \begin{tabular}{lccc}
    \Xhline{2\arrayrulewidth}
    NFE & 10   & 25   & 50   \\ \hline\hline
    DDIM         & 18.85 & 9.79 & 7.17 \\
    D-DDIM          & \textbf{8.67} & 8.18 & 6.55 \\
    Fixed-D-DDIM ($\lambda=0.5$)   & 11.45 & \textbf{7.00} & \textbf{5.27} \\
    LA-DDIM  ($\lambda=0.1$)  & 15.24 & 8.57 & 6.29 \\ \Xhline{2\arrayrulewidth}
    \end{tabular}
    \vspace{0.5em}
     \caption{\textbf{Comparison with learning-free samplers} on CIFAR-10 with noise prediction models. The best FID is bolded.}
    \label{suptab:comparison_la}
\end{table}

In this section, we present further comparisons between D-ODE solvers and previous learning-based (knowledge distillation) and learning-free methods. \cref{supfigA:comparison_distillation} displays FID scores with varying NFE on CIFAR-10, including consistency distillation (CD)~\citep{song2023consistency}, which can perform a one-step or few-step sampling, and progressive distillation (PD)~\citep{salimans2022progressive}, allowing a sampling with steps in a geometric sequence (\eg, 1, 2, 4, ..., 1024). D-EDM requires at least two steps to utilize previous denoising outputs. 

Overall, CD outperforms other methods in terms of FID on one-step generation. However, it is important to note that this comparison does not account for training time. For instance, \citet{song2023consistency} reported that consistency models on CIFAR-10 utilized 8 Nvidia A100 GPUs for training. On the other hand, simply generating 50K samples for 30 steps takes less than 30 minutes on a single A100 GPU, achieving similar sample quality to consistency models. while CD and PD are attractive options for practitioners with ample computational resources, given their ability to enable one-step generation, the major advantage of D-ODE solvers lies in their capacity to enhance existing ODE solver-based samplers with minimal modifications and fast optimization.

Recently, \citet{zhang2023lookahead} introduced lookahead diffusion models which enhance the FID scores of existing ODE solvers by refining mean estimation using previous data predictions. They achieve this by extrapolating previous predictions of initial data to approximate the target data. Unlike D-ODE solvers, lookahead models require parameter $\lambda$ to be chosen through grid search, with a default setting of $\lambda=0.1$ during experiments. Following their configuration, we compare lookahead diffusion models of DDIM, so-called LA-DDIM, with our D-DDIM in \cref{suptab:comparison_la}. The table shows that, except at 50 NFE, D-DDIM outperforms LA-DDIM.

Inspired by LA-DDIM, we also experiment with fixing $\lambda_t$ in D-DDIM as a constant $\lambda$ and optimizing it through grid search. We refer to this modified approach as Fixed-D-DDIM. In \cref{supfigB:comparison_LA} and \cref{supfigC:comparison_Fixed_d_ddim}, we conduct grid searches on $\lambda$ using a 10-step sampler on CIFAR-10. Additionally, we provide the FID scores of DDIM and D-DDIM as references (dotted lines). Despite the grid search performed on LA-DDIM, it is unable to match the FID of D-DDIM. On the other hand, Fixed-D-DDIM achieves the same FID as D-DDIM with sufficient grid search.  This suggests that leveraging denoising outputs is a more efficient strategy than relying on initial data predictions. Moreover, Fixed-D-DDIM further improves upon D-DDIM's performance at 25 and 50 NFE, indicating the potential for finding an even better $\lambda$ value that results in a lower FID. Future research directions could explore various methods to efficiently determine $\lambda$. It is important to highlight that the FID of LA-DDIM and Fixed-D-DDIM varies depending on the chosen $\lambda$. However, D-DDIM's advantage over other methods is its independence from grid search, with sampling times comparable to DDIM.

\section{More Experiments on DPM-Solver++}

\begin{figure*}[h]
	\centering
	\begin{subfigure}[t]{0.3\linewidth}
	        \includegraphics[width=\linewidth]{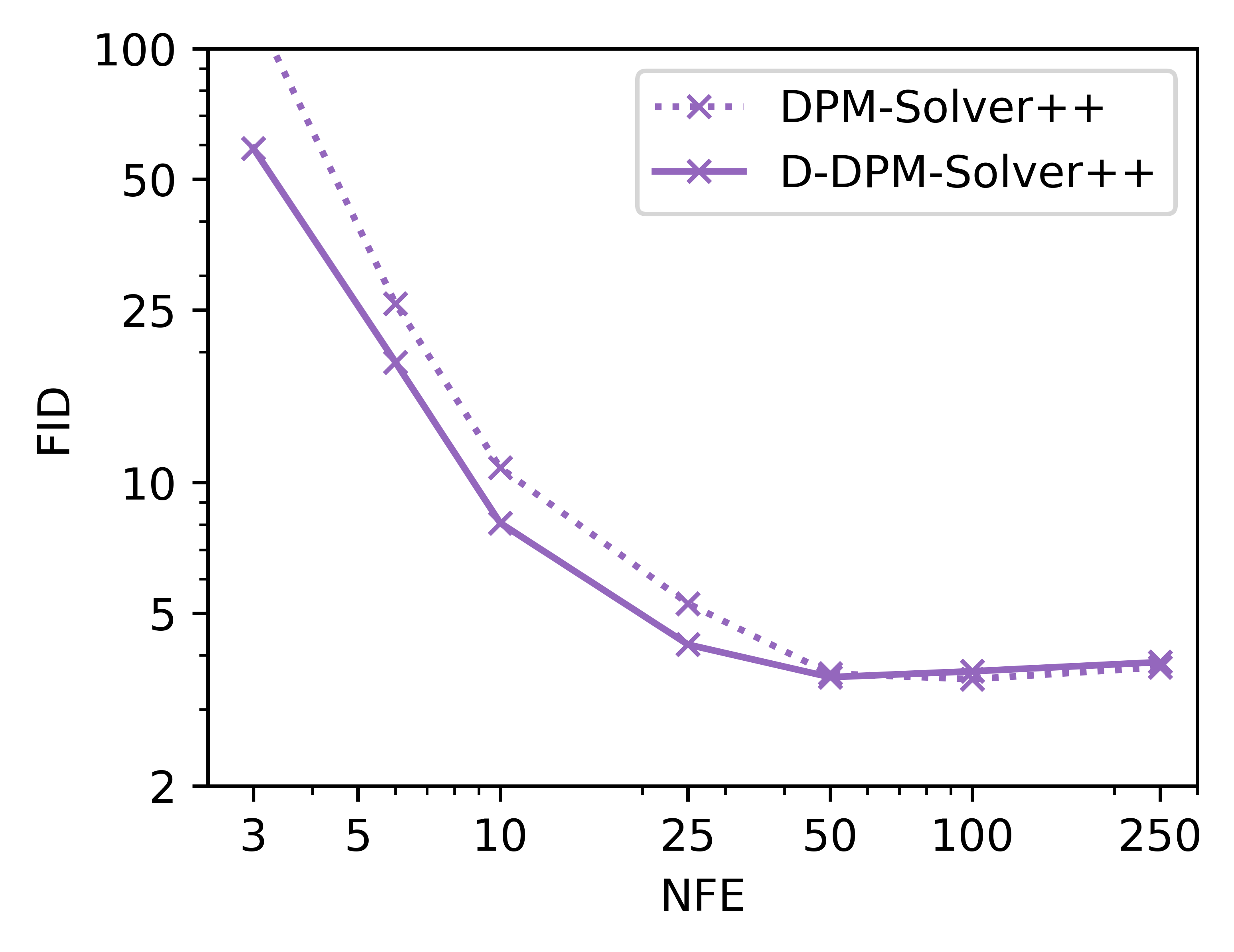}
	        \caption{CIFAR-10 $(32\times32)$}
            \label{supfigA:cifar10_dpmsolver++}
         \end{subfigure}
	\begin{subfigure}[t]{0.3\linewidth}
		\includegraphics[width=\linewidth]{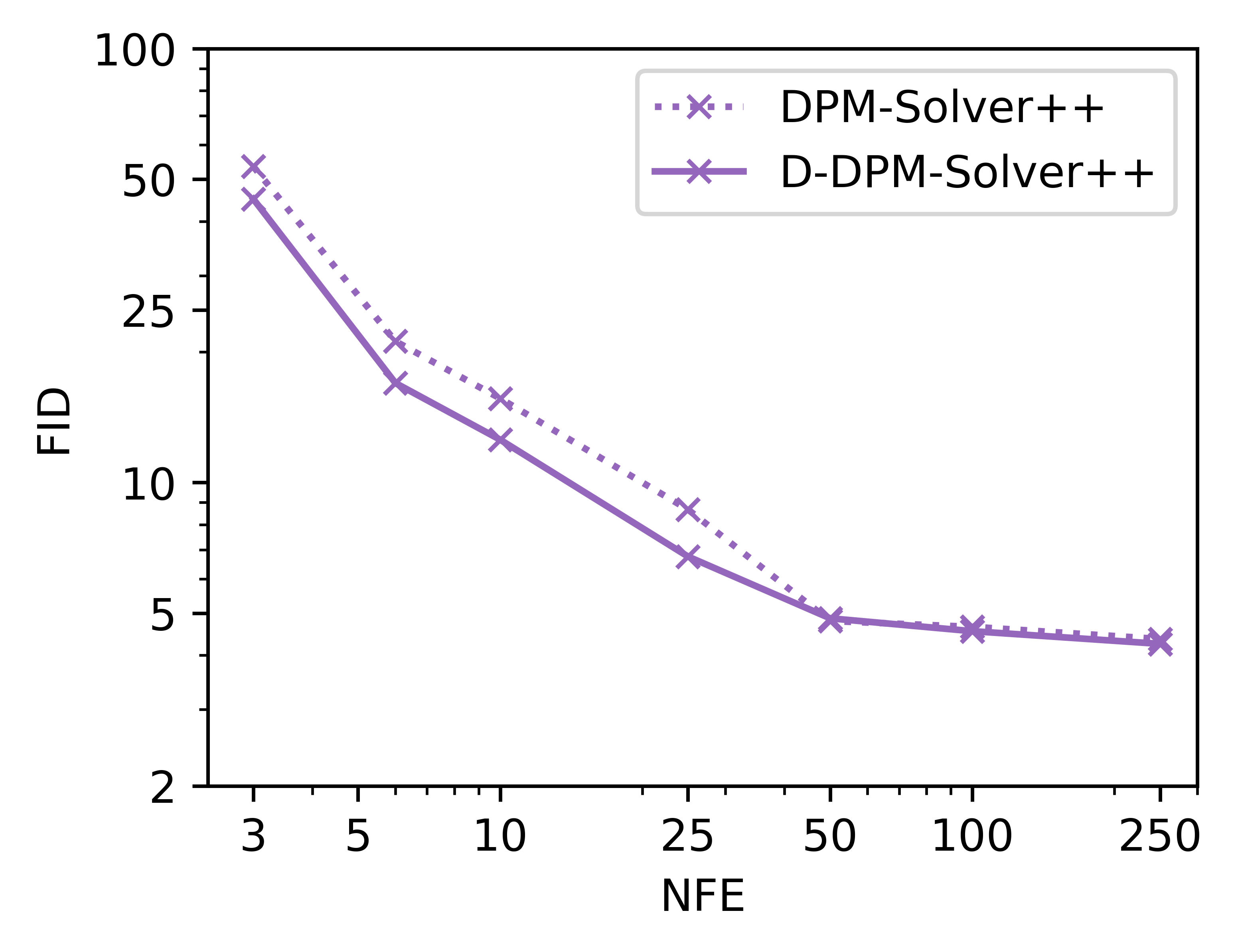}
		\caption{CelebA $(64\times64)$}
         \label{supfigB:celeba_dpmsolver++}
	\end{subfigure}
	\begin{subfigure}[t]{0.3\linewidth}
		\includegraphics[width=\linewidth]{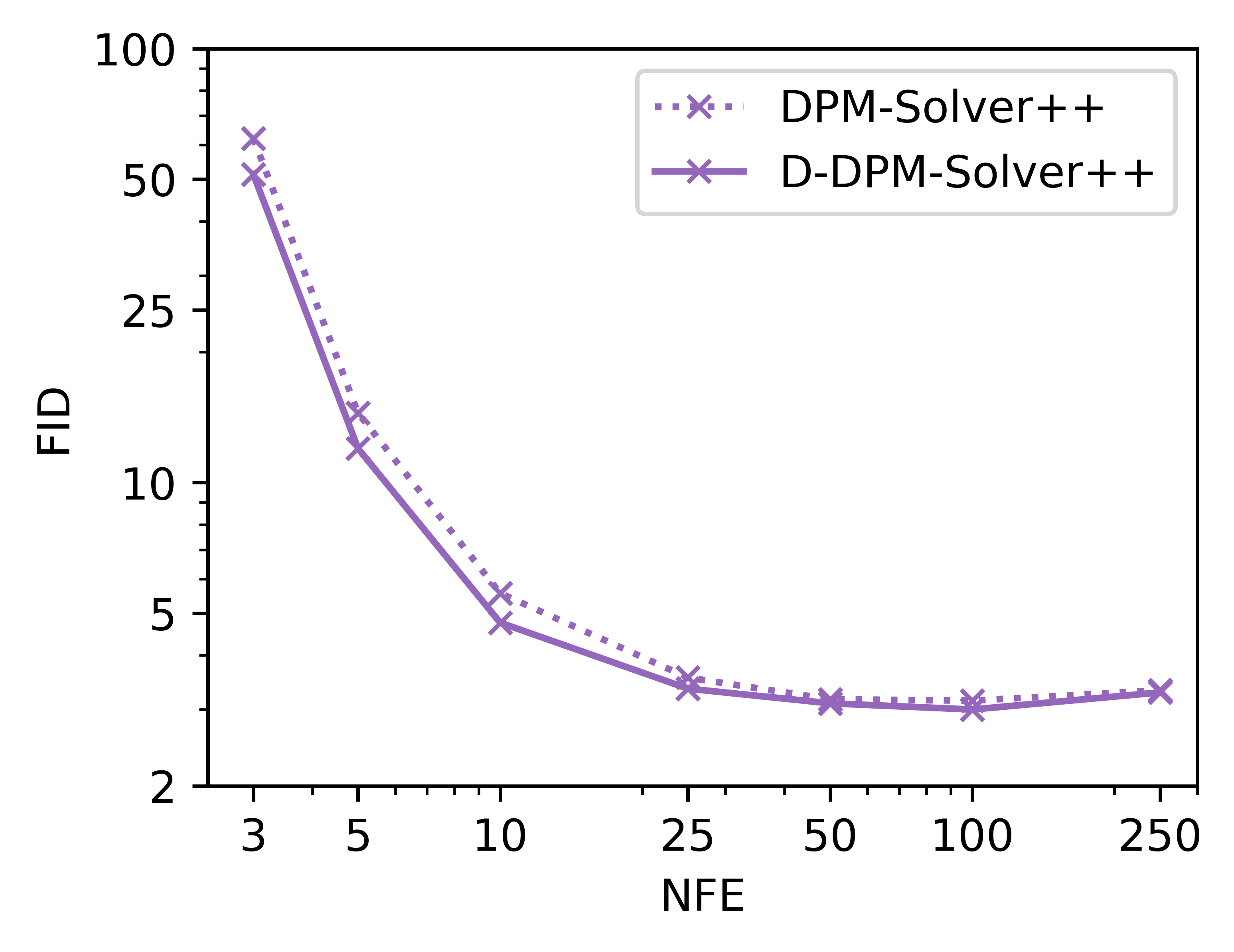}
	\caption{ImageNet $(128\times128)$}
        \label{supfigC:imagenet_dpmsolver++}
	\end{subfigure}
	\caption{\textbf{Image quality measured by FID $\downarrow$ with DPM-Solver++.} Dotted lines denote DPM-Solver++ while straight lines represent D-DPM-Solver++.}
 \label{supfig:dpmsolver++}
\end{figure*}

\begin{figure*}[h]
	\centering
	\begin{subfigure}[t]{\linewidth}
		\includegraphics[width=\linewidth]{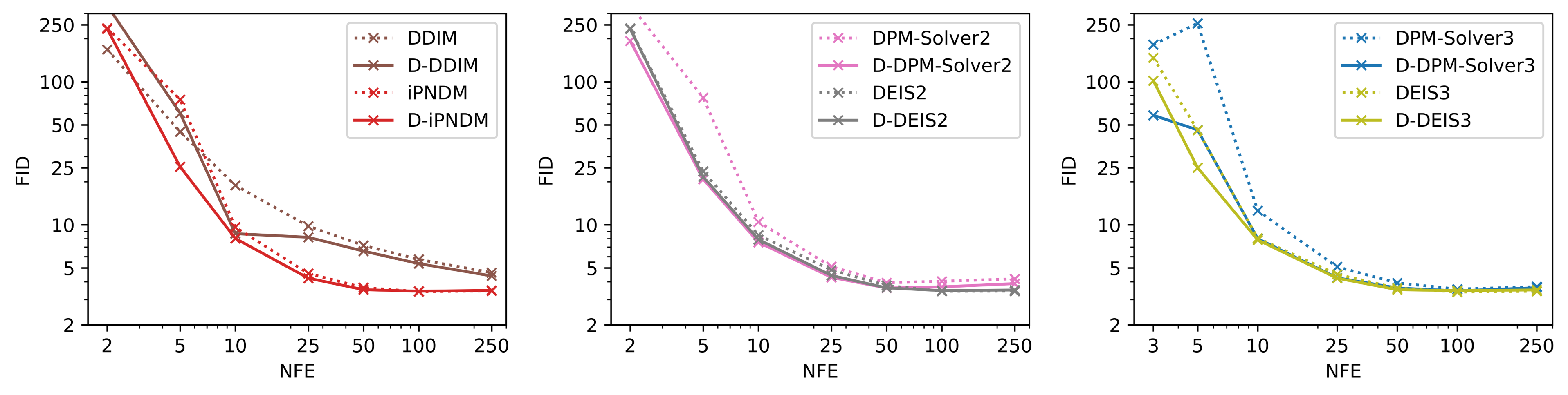}
		\caption{CIFAR-10 $(32\times32)$}
		\label{supfig:noise_subfigA}
	\end{subfigure} \\
        
	\begin{subfigure}[t]{\linewidth}
		\includegraphics[width=\linewidth]{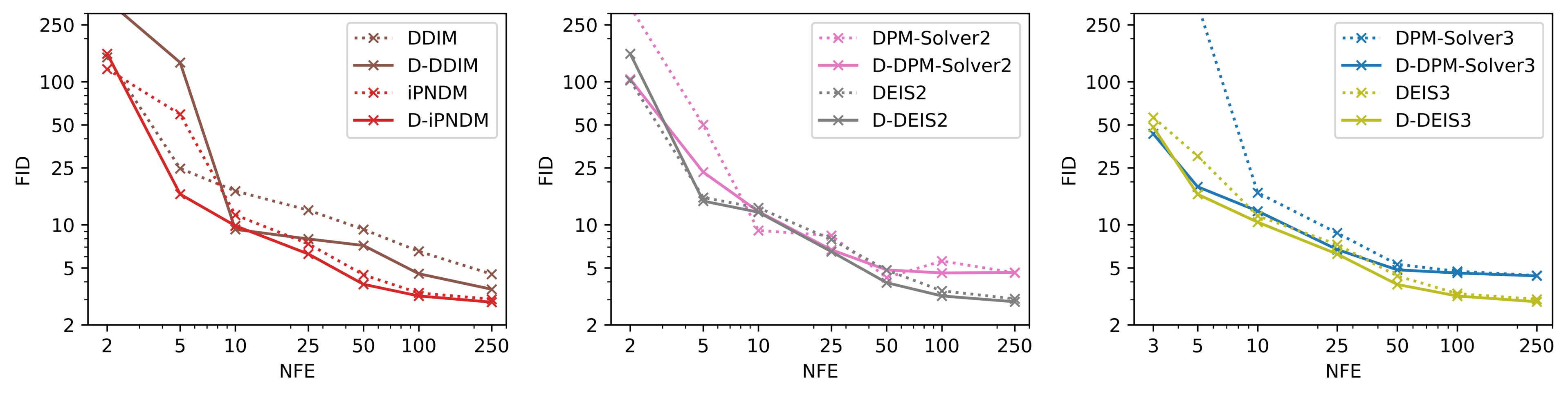}
		\caption{CelebA $(64\times64)$}
		\label{supfig:noise_subfigB}
	\end{subfigure} \\

	\begin{subfigure}[t]{\linewidth}
            \includegraphics[width=\linewidth]{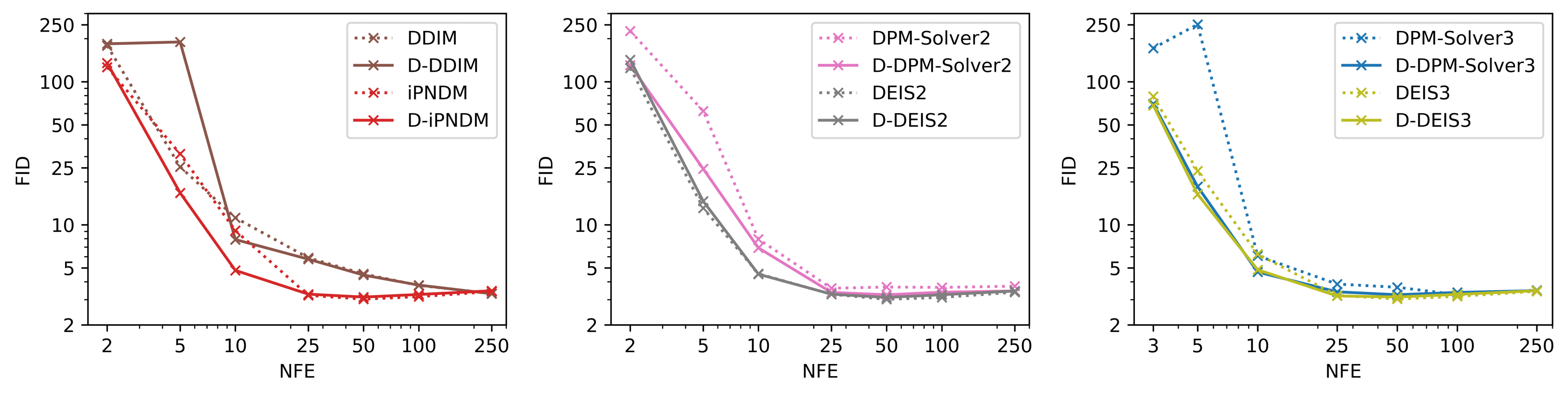}
            \caption{ImageNet $(128\times128)$}
            \label{supfig:noise_subfigC}
        \end{subfigure} 
	\caption{\textbf{Image quality measured by FID $\downarrow$} with varying NFE $\in \{2, 5, 10, 25, 50, 100, 250\}$. For DPM-Solver3 and DEIS3, we use 3 NFE instead of 2 NFE as the third-order method requires at least three denoising outputs. Dotted lines denote ODE solvers (DDIM, iPNDM, DPM-Solver, and DEIS) while straight lines represent the applications of D-ODE solver to them (D-DDIM, D-iPNDM, D-DPM-Solver, and D-DEIS).}
	\label{supfig:noise_subfigures}
\end{figure*}

Built upon DPM-Solver~\cite{lu2022dpm}, DPM-Solver++~\cite{lu2022dpm++} addresses the instability in the previous multi-step approach of solving diffusion ODE and adopts thresholding methods to constrain the solution within the range of the original data. Similar to the formulation of D-DPM-Solver explained in \cref{sup_subsec:d_ode_noise}, we apply our new denoising outputs to replace the original denoising output. \cref{supfig:dpmsolver++} demonstrates that applying D-ODE solvers to DPM-Solver++ can further improve the image quality through distillation.

In addition, we present extra experiment results in \cref{supfig:noise_subfigures} with noise prediction models on CIFAR-10, CelebA64, and ImageNet128.

\section{Analysis Figures and Qualitative Results}~\label{sup:extra_experiments}

\begin{figure*}[h]
\begin{center}
    \includegraphics[width=1.0\linewidth]{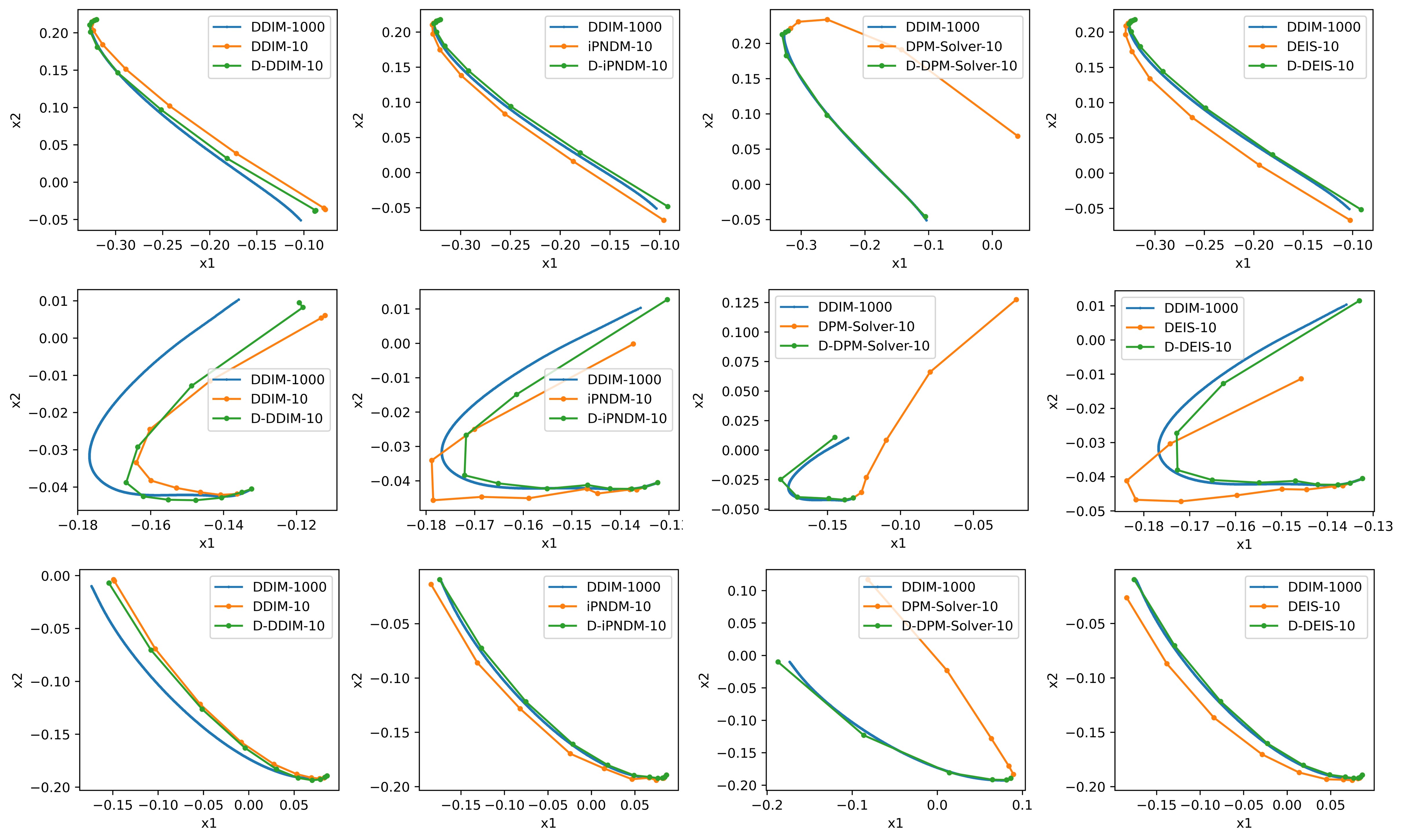}
\end{center}
\caption{\textbf{Update path of randomly selected two pixels in the images.} The result of 1000-step DDIM is displayed as our target. These figures are drawn with 1000 samples using a noise prediction model trained on CIFAR-10.}
\label{supfig:more_visual_analysis}
\end{figure*}

\begin{figure*}[h]
	\centering
	\begin{subfigure}[t]{0.78\linewidth}
		\includegraphics[width=\linewidth]{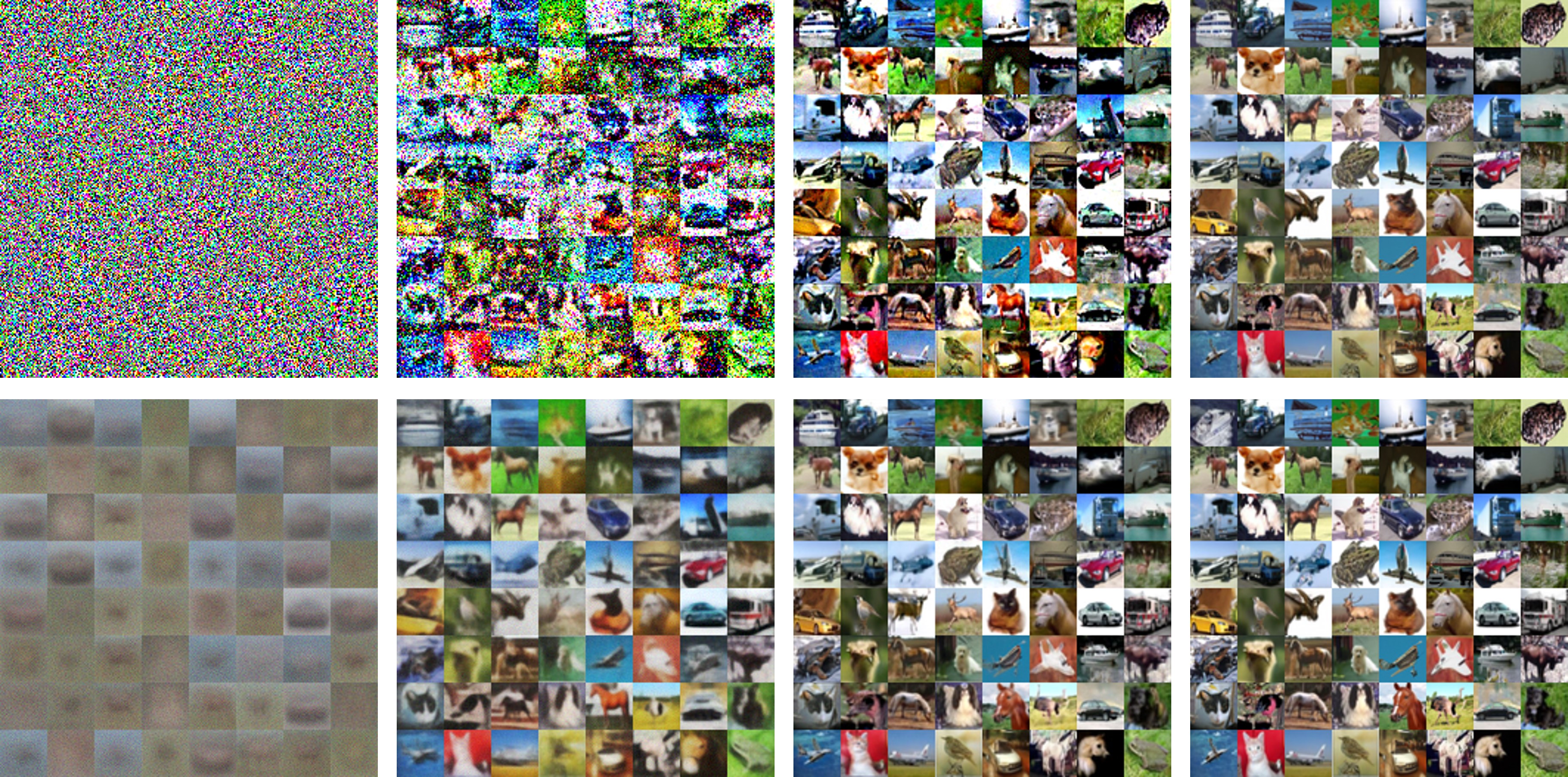}
		\caption{CIFAR-10 $(32\times32)$}
	\end{subfigure} \\
        
	\begin{subfigure}[t]{0.78\linewidth}
		\includegraphics[width=\linewidth]{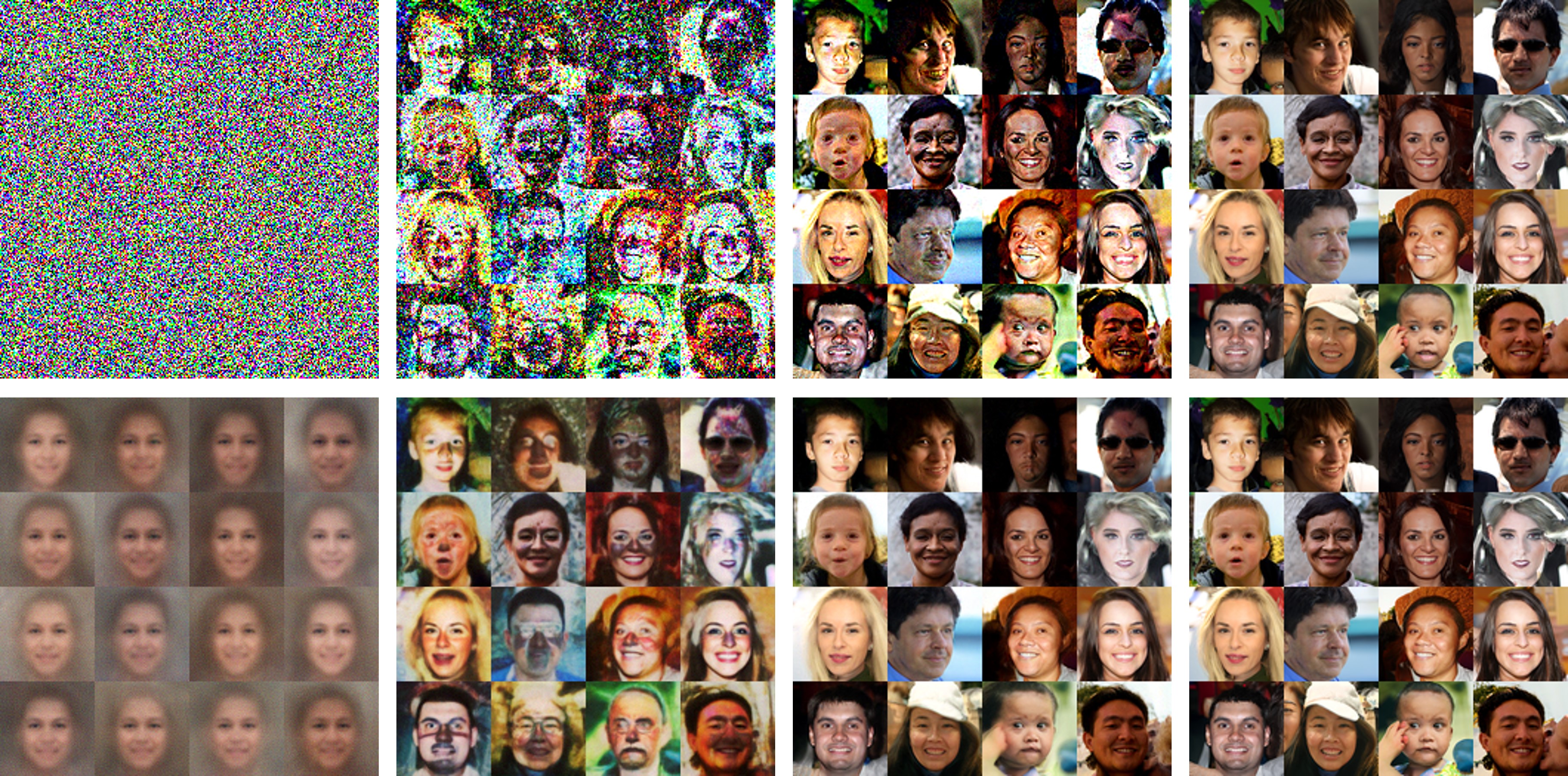}
		\caption{FFHQ $(64\times64)$}
	\end{subfigure} \\
 
	\begin{subfigure}[t]{0.78\linewidth}
		\includegraphics[width=\linewidth]{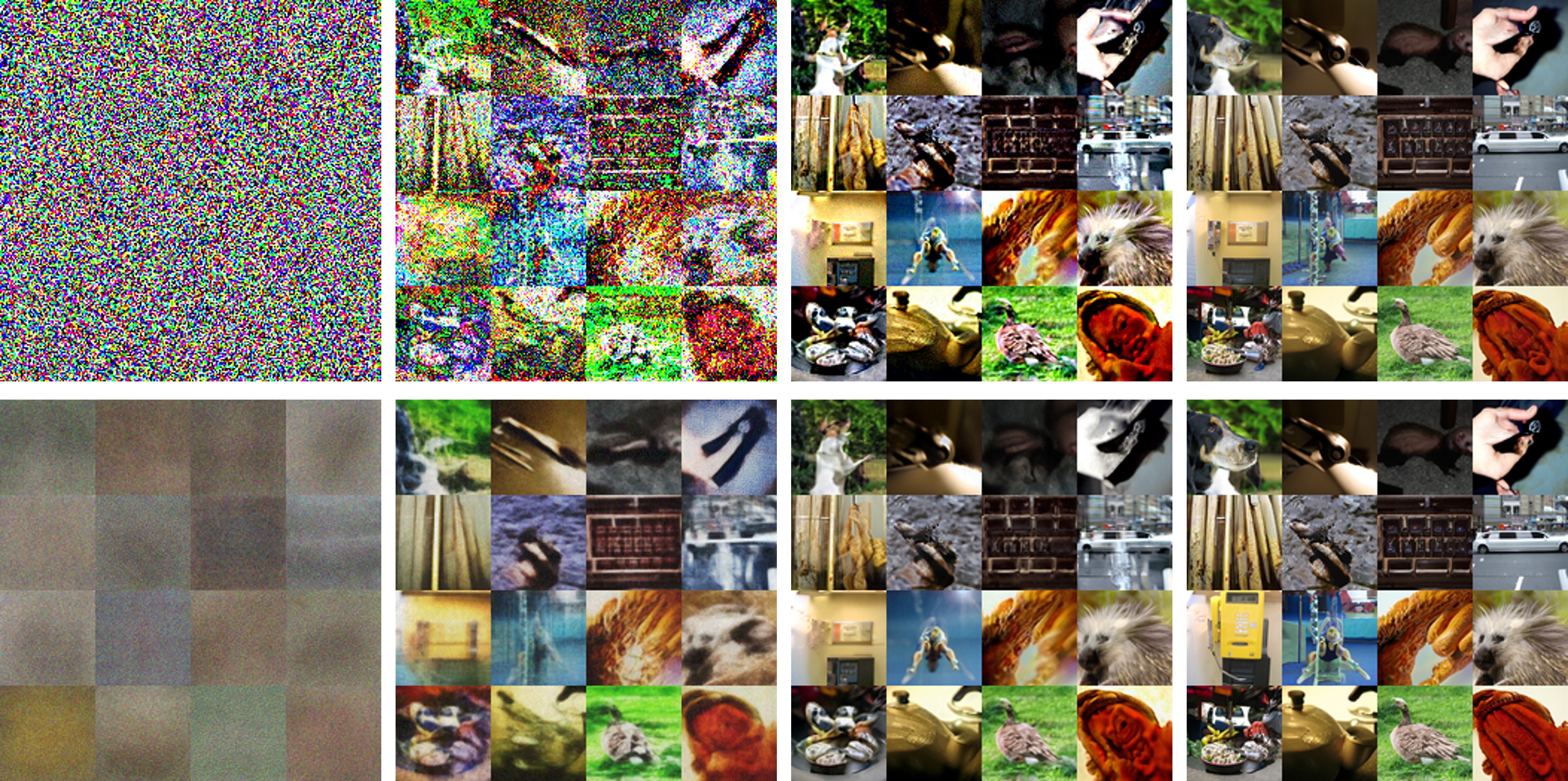}
		\caption{ImageNet $(64\times64)$}
	\end{subfigure}
\caption{\textbf{Qualitative results of CIFAR-10 $(32\times32)$, FFHQ $(64\times64)$, and ImageNet $(64\times64)$ with data prediction models.} We compare EDM (top) and D-EDM (bottom) in each subfigure with NFE $\in \{3, 5, 9, 25\}$.}
\label{supfig:qualtiative_results1}
\end{figure*}

\begin{figure*}[h]
	\centering
	\begin{subfigure}[t]{\linewidth}
		\includegraphics[width=\linewidth]{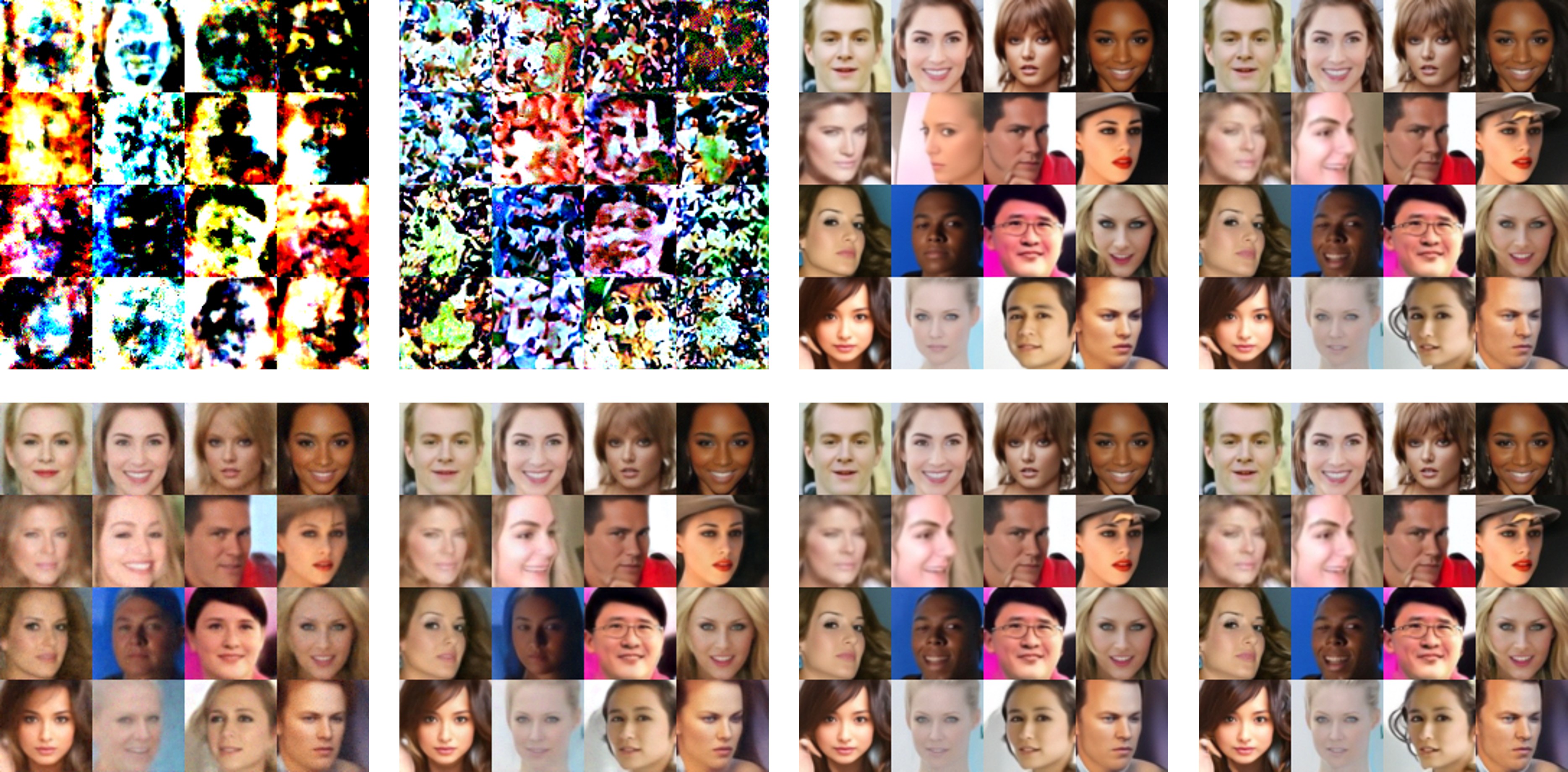}
		\caption{DPM-Solver3 (top) vs D-DPM-Solver3 (bottom)}
	\end{subfigure} \\
        
	\begin{subfigure}[t]{\linewidth}
		\includegraphics[width=\linewidth]{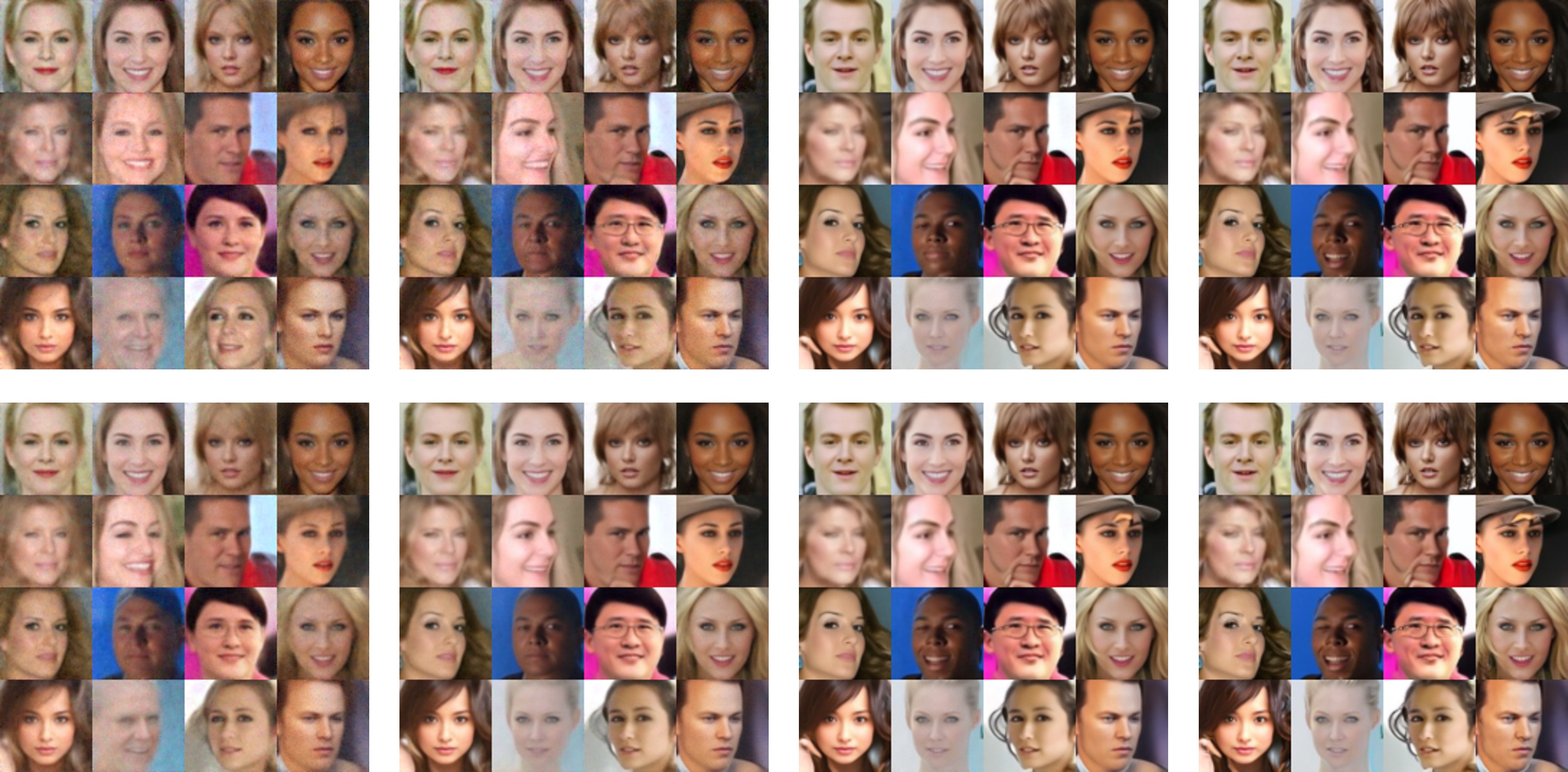}
		\caption{DEIS3 (top) vs D-DEIS3 (bottom)}
	\end{subfigure} 
\caption{\textbf{Qualitative results of CelebA $(64\times64)$ with noise prediction models.} We compare ODE-solvers (DPM-Solver3, DEIS3) and D-ODE solvers (D-DPM-Solver3, D-DEIS3) in each subfigure with NFE $\in \{3, 5, 10, 25\}$.}
\label{supfig:qualtiative_results3}
\end{figure*}

\begin{figure*}[h]
	\centering
	\begin{subfigure}[t]{\linewidth}
		\includegraphics[width=\linewidth]{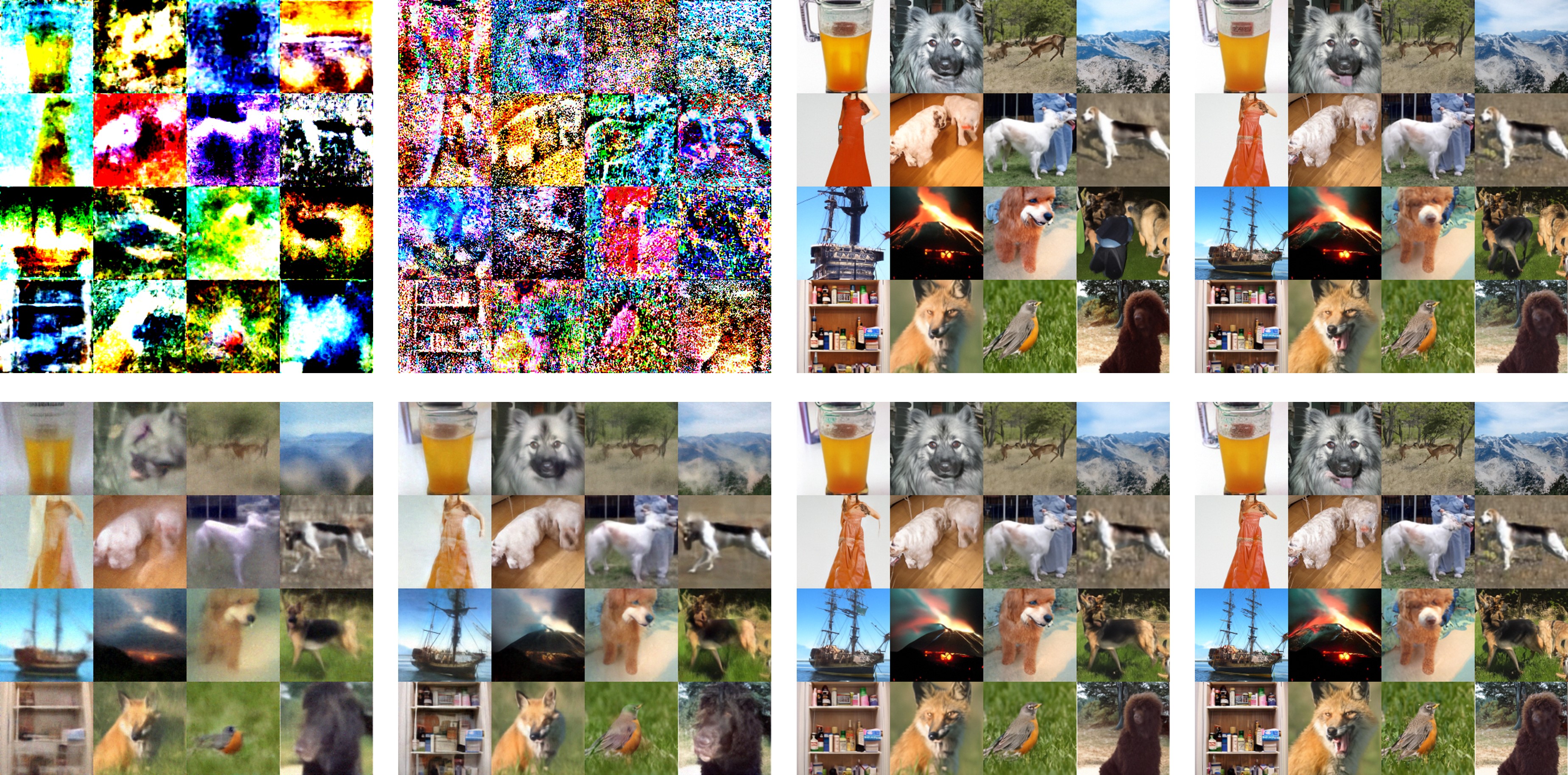}
		\caption{DPM-Solver3 (top) vs D-DPM-Solver3 (bottom)}
	\end{subfigure} \\
        
	\begin{subfigure}[t]{\linewidth}
		\includegraphics[width=\linewidth]{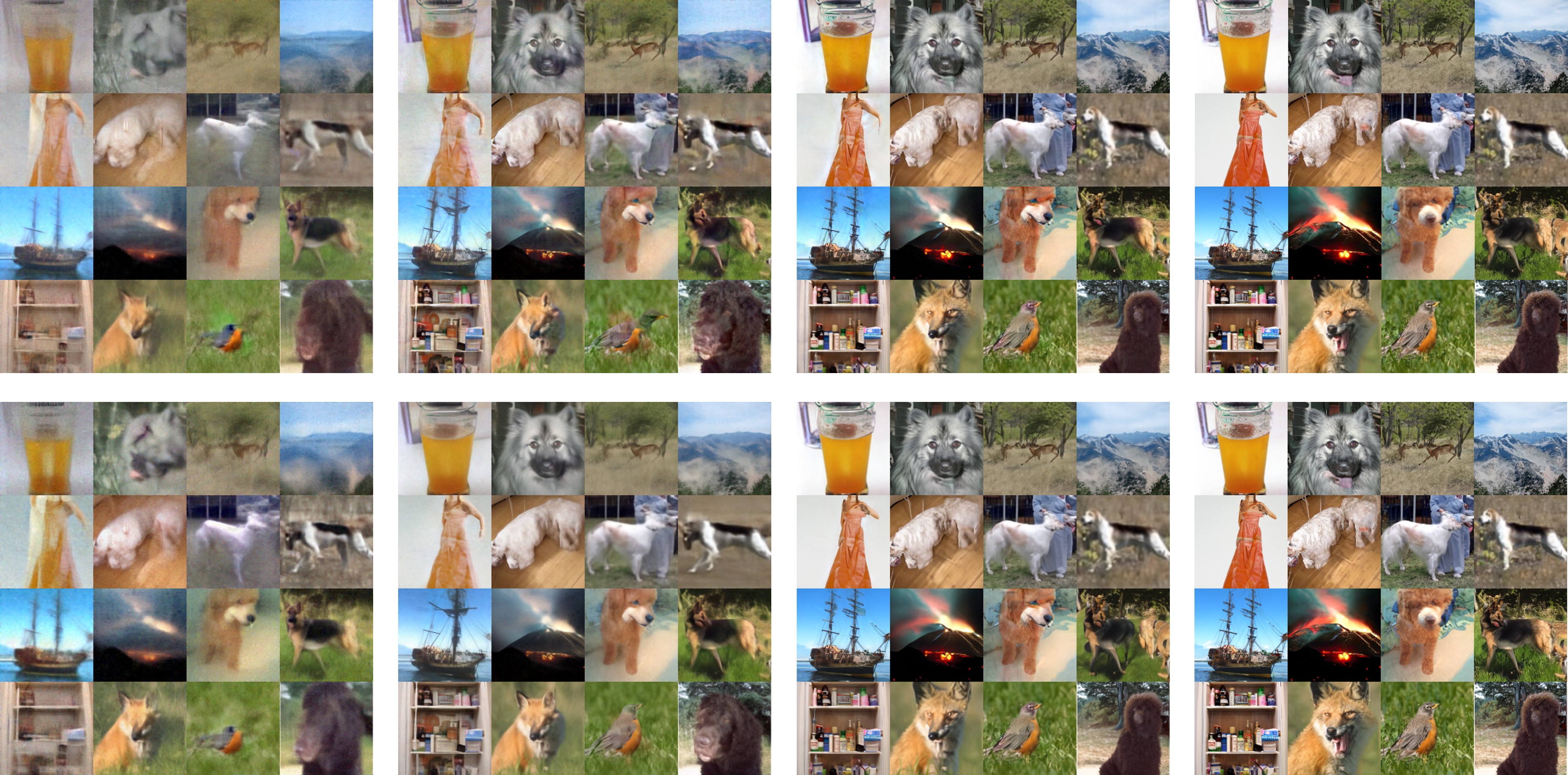}
		\caption{DEIS3 (top) vs D-DEIS3 (bottom)}
	\end{subfigure} 
\caption{\textbf{Qualitative results of ImageNet $(128\times128)$ with noise prediction models.} We compare ODE-solvers (DPM-Solver3, DEIS3) and D-ODE solvers (D-DPM-Solver3, D-DEIS3) in each subfigure with NFE $\in \{3, 5, 10, 25\}$.}
\label{supfig:qualtiative_results4}
\end{figure*}

\begin{figure*}[h]
	\centering
	\begin{subfigure}[t]{\linewidth}
		\includegraphics[width=\linewidth]{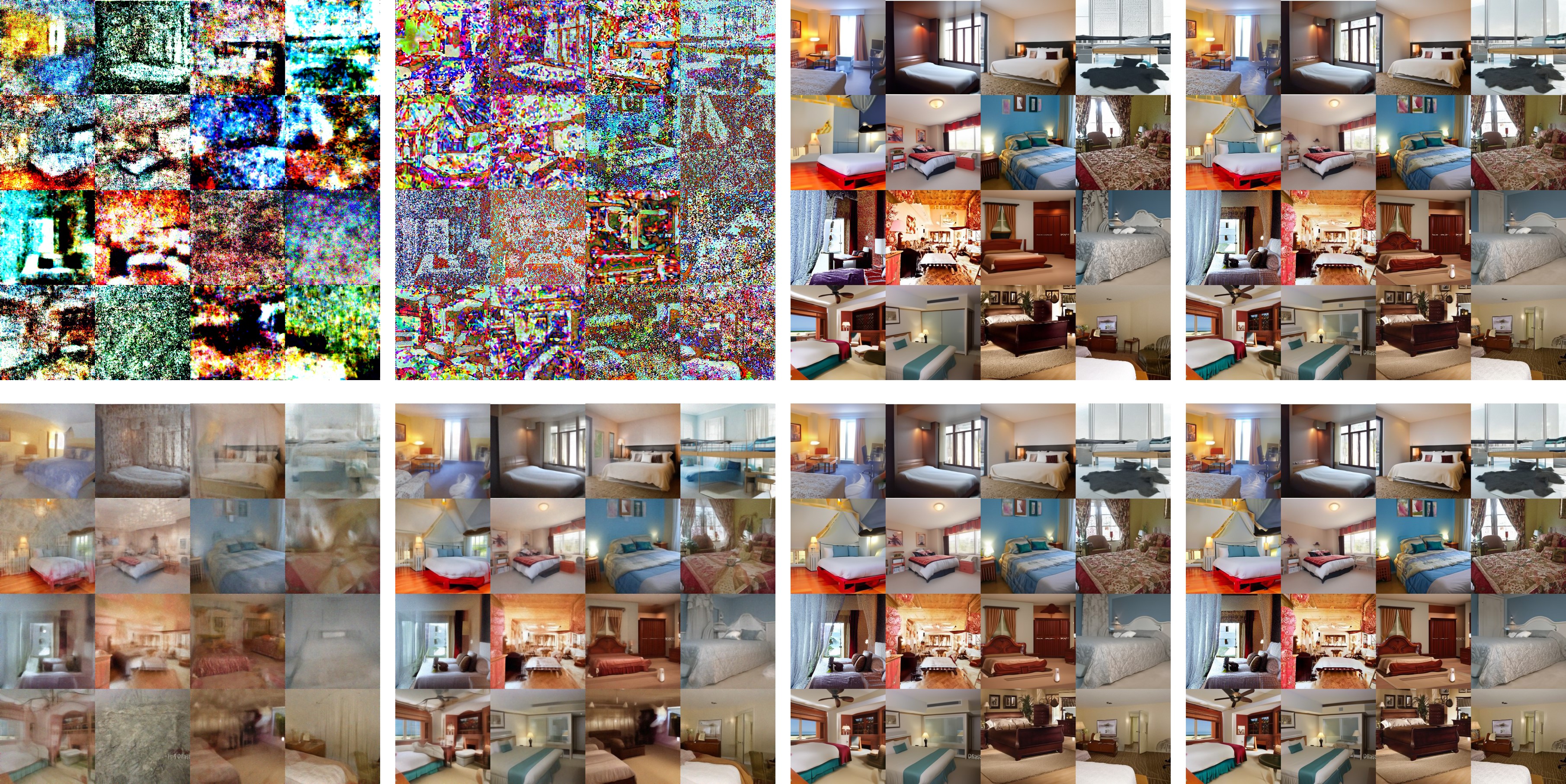}
		\caption{DPM-Solver3 (top) vs D-DPM-Solver3 (bottom)}
	\end{subfigure} \\
        
	\begin{subfigure}[t]{\linewidth}
		\includegraphics[width=\linewidth]{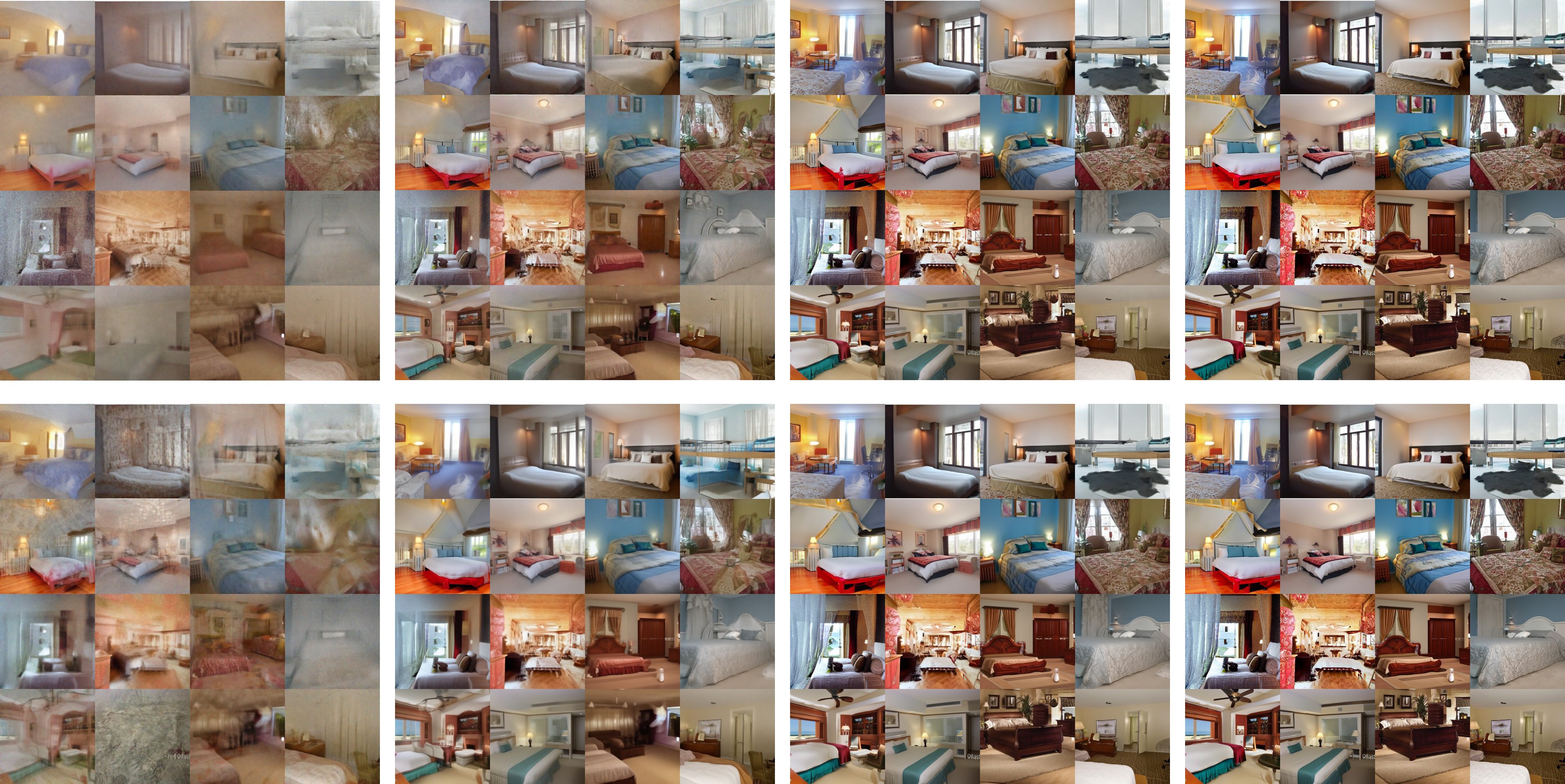}
		\caption{DEIS3 (top) vs D-DEIS3 (bottom)}
	\end{subfigure} 
\caption{\textbf{Qualitative results of LSUN Bedroom $(256\times256)$ with noise prediction models.} We compare ODE-solvers (DPM-Solver3, DEIS3) and D-ODE solvers (D-DPM-Solver3, D-DEIS3) in each subfigure with NFE $\in \{3, 5, 10, 25\}$.}
\label{supfig:qualtiative_results5}
\end{figure*}

In \cref{supfig:more_visual_analysis}, more analysis figures similar to Fig.\ 5 of the main paper are shown with different pixels. We also show more qualitative results in \cref{supfig:qualtiative_results1}, \cref{supfig:qualtiative_results3}, \cref{supfig:qualtiative_results4}, and \cref{supfig:qualtiative_results5}.

\clearpage

\newpage
{
    \small
    \bibliographystyle{ieeenat_fullname}
    \bibliography{main}
}


\end{document}